\documentclass[journal]{IEEEtran}
\usepackage{amsmath,amsfonts}
\usepackage{algpseudocode}
\usepackage{algorithm}
\usepackage{array}
\usepackage{textcomp}
\usepackage{stfloats}
\usepackage{verbatim}
\usepackage{graphicx}
\usepackage{cite}
\usepackage{color}
\usepackage{bm}
\usepackage{makecell}
\usepackage{tabularx}
\usepackage{subfigure}
\usepackage{csquotes}
\usepackage{amsthm}
\usepackage{makecell}
\usepackage{booktabs}
\usepackage{hyperref}
\usepackage{multirow}
\usepackage{multicol}
\usepackage[normalem]{ulem}
\usepackage{tablefootnote}
\algrenewcommand\algorithmicrequire{\textbf{Input:}}
\algrenewcommand\algorithmicensure{\textbf{Output:}}
\algrenewcommand\algorithmicindent{1.0em}
\begin{document}
\title{Effects of Archive Size on Computation Time and Solution Quality for Multi-Objective Optimization}

\author{Tianye Shu, Ke Shang, Hisao Ishibuchi,~\IEEEmembership{Fellow,~IEEE}, Yang Nan
\thanks{This work was supported by National Natural Science Foundation of China (Grant No. 62002152, 61876075), Guangdong Provincial Key Laboratory (Grant No. 2020B121201001), the Program for Guangdong Introducing Innovative and Enterpreneurial Teams (Grant No. 2017ZT07X386), The Stable Support Plan Program of Shenzhen Natural Science Fund (Grant No. 20200925174447003), Shenzhen Science and Technology Program (Grant No. KQTD2016112514355531). \textit{(Corresponding Author: Ke Shang and Hisao Ishibuchi.)}}
\thanks{T. Shu, K. Shang, H. Ishibuchi, Y. Nan are with Guangdong Provincial Key Laboratory of Brain-inspired Intelligent Computation, Department of Computer Science and Engineering, Southern University of Science and Technology, Shenzhen 518055, China (e-mail: 12132356@mail.sustech.edu.cn; kshang@foxmail.com; hisao@sustech.edu.cn; nany@mail.sustech.edu.cn).}
}



\maketitle
\begin{abstract}
\textcolor{black}{An unbounded external archive has been used to store all nondominated solutions found by an evolutionary multi-objective optimization algorithm in some studies. It has been shown that a selected solution subset from the stored solutions is often better than the final population.} However, the use of the unbounded archive is not always realistic. When the number of examined solutions is huge, we must pre-specify the archive size. In this study, we examine the effects of the archive size on three aspects: (i) the quality of the selected final solution set, (ii) the total computation time for the archive maintenance and the final solution set selection, and (iii) the required memory size. \textcolor{black}{Unsurprisingly, the increase of the archive size improves the final solution set quality. Interestingly, the total computation time of a medium-size archive is much larger than that of a small-size archive and a huge-size archive (e.g., an unbounded archive).} To decrease the computation time, we examine two ideas: periodical archive update and archiving only in later generations. \textcolor{black}{Compared with 
updating the archive at every generation,} the first idea can obtain almost the same final solution set quality using a much shorter computation time at the cost of a slight increase of the memory size. The second idea drastically decreases the computation time at the cost of a slight deterioration of the final solution set quality. Based on our experimental results, some suggestions are given about how to appropriately choose an archiving strategy and an archive size.
\end{abstract}

\begin{IEEEkeywords}
Evolutionary multi-objective optimization, evolutionary many-objective optimization, external archive, solution subset selection, performance evaluation.
\end{IEEEkeywords}
\section{Introduction}
In the evolutionary multi-objective optimization (EMO) field, an EMO algorithm is used to search for a set of nondominated solutions which approximates the entire Pareto front of a multi-objective optimization problem (MOP). 
\textcolor{black}{When an MOP has $M$ conflicting and continuous objectives, its Pareto front forms an $(M-1)$-dimensional manifold~\cite{hillermeier2001nonlinear,schutze2008hybr}. Thus, we usually need a large number of nondominated solutions to approximate the entire Pareto front.} To store those solutions, various archiving strategies have been proposed~\cite{schutze2021archiving}. Many EMO algorithms implicitly use the current population as an archive and present the final population as a final solution set to the decision marker (e.g., NSGA-II~\cite{deb2002fast}). \textcolor{black}{Some other EMO algorithms (e.g., algorithms in Table~\ref{tab:archiveSizeTable}) have an external archive and update it to store a pre-specified number of nondominated solutions. The external archive is presented to the decision marker as a final solution set after the termination of those algorithms. Therefore, the archive sizes of those algorithms are not large as shown in Table~\ref{tab:archiveSizeTable}.}\par
\begin{table}[!b]
    \vspace{-2em}
\centering
    \caption{
        Specifications of the population size and the archive size in EMO algorithms with external archives in the literature}
    \begin{tabular}{|m{1.34cm}|m{3.4cm}|m{2.65cm}|}\hline
    Algorithm & Population Size& Archive Size\\ \hline
    SPEA2~\cite{zitzler2001spea2} & 250, 300 and 400 for knapsack problems with 2, 3 and 4 objectives.\newline 100 for continuous problems. & Same as the population size \\\hline
      \makecell[l]{Two\_Arch\\~\cite{Pradi2006anew}}& 20, 50, 100, 250 and 600 for 2, 3, 4, 6 and 8 objectives, respectively.\newline The setting was based on Khare \emph{et al.}~\cite{khare2003performance}.&Same as the population size\\ \hline \makecell[l]{Two\_Arch2\\~\cite{wang2015two_arch2}} &Convergence archive$^a$:\par\;\;10, 50, 100, 200 and 300 are tested on DTLZ1 with 10 objectives, and 100 is recommended.&Diversity archive:\par\;\;200 for problems with 11 or more objectives\par\;\;100 for problems with 10 or less objectives \\\hline
      EAG-\newline MOEA/D~\cite{cai2015external} &100 for one problem\newline 200 for another problem &Same as the population size\\\hline
      AMGA~\cite{tiwari2008amga}&100 for the initial population\newline 8 for the parent population &100 \\\hline
      AMGA2~\cite{tiwari2011amga2}&100 for the initial population\newline $4M$ for the parent population\newline ($M$ is the number of objectives)& 100\\\hline
    \end{tabular}
     \vspace{0.2em}
   {\footnotesize{
  
    $^a$ Here the convergence archive is viewed as the main population since the mutation is applied to only solutions in the convergence archive and the diversity archive is presented to the decision maker in Two\_Arch2.}}
    \label{tab:archiveSizeTable}
\end{table}
\textcolor{black}{Compared with an unbounded archive, a small-size archive can be updated more efficiently. However, a bounded population (archive) often leads to shrinking and oscillation problems and degrades the performance of the final solution sets~\cite{fieldsend2003using}. Therefore, an unbounded external archive (UEA) is used in some studies~\cite{Ishibuchi1998Amulti,parks1998selective,fieldsend2003using,schutze2008hybr} to store all examined solutions.} Recently, some studies~\cite{bringmann2014generic, ishibuchi2016compare, ishibuchi2020new} showed that the solution sets which are better than the final population (archive) can be found by subset selection from all solutions examined during the execution of an EMO algorithm. This is because good solutions can be deleted from a bounded population (archive) before the final generation. It was also shown in~\cite{li2019empirical} that the final population (archive) often includes solutions which are dominated by other solutions generated and discarded in earlier generations. Thus, it is a good idea to select a pre-specified number of solutions from all examined solutions stored in an unbounded external archive. Tanabe et al.~\cite{tanabe2017benchmarking} compared the performance of different EMO algorithms using selected solutions from unbounded archives. Bezerra et al.~\cite{bezerra2019archiver} demonstrated that selected solutions from bounded and unbounded archives are better than the final population. Based on these results, the use of an EMO algorithm framework with an unbounded archive was proposed in~\cite{ishibuchi2020new} for the design of new EMO algorithms. All of these studies showed the usefulness of the unbounded archive.\par
In some real-world applications, solution evaluation is expensive, which means that the number of examined solutions is small. In this case, it is easy to store all examined solutions. In most real-world applications, storing the evaluation result of a solution is much cheaper than its evaluation. Thus, the use of an unbounded archive in an EMO algorithm is usually realistic. However, when the solution evaluation is not expensive and the search for the Pareto front is very difficult, a huge number of solutions are examined in a single run of an EMO algorithm. For example, the total number of examined solutions was specified as 230 million for large-scale many-objective optimization in some experiments in~\cite{zhang2018decision}. \textcolor{black}{In this case, it is unrealistic to store all examined solutions. Thus, we need to pre-specify the archive size (and use a truncation operation to remove excess solutions).}\par
As we have already explained, the performance of existing EMO algorithms can be improved by selecting a final solution set from an archive. However, the effect of the archive size has not been well studied in the literature. A large archive has not been examined in EMO algorithms (except for the above-mentioned studies on subset selection from all examined solutions). In this paper, we examine the effects of the archive size on the following three aspects: (i) the quality of a selected final solution set, (ii) the total computation time for the archive maintenance and the final solution set selection, and (iii) the required memory size for the archive. It is shown that the increase in the archive size improves the quality of the selected final solution set. The total computation time severely increases with the increase in the archive size. However, after reaching its maximum value, it starts to decrease. \textcolor{black}{That is, the computation time is small both when the archive size is small and when the archive size is huge (e.g., similar to the total number of examined solutions)}. This means that it is more time-efficient to select a final solution set from all examined solutions than to maintain a medium-size archive (e.g., an archive with $50N$ solutions where $N$ is the population size). To reduce the computation time for archiving, we propose two strategies. One is a lazy periodical strategy where the archive is updated periodically (i.e., at every $T$ generations). This strategy is to decrease the computation time without deteriorating the quality of the final solution set at the cost of a slight increase of the memory size. The other is a last $X$-generation strategy where all solutions only in the last $X$ generations are stored. This strategy is to drastically decrease the computation time without increasing the memory size at the cost of a slight deterioration of the final solution set quality. Our experiments are performed using different archiving strategies with various archive size specifications. Based on experimental results, some suggestions are given about how to choose an appropriate archiving strategy and an appropriate archive size.\par
\section{Background}\label{sec:Background}
\subsection{Subset Selection}
In general, the subset selection problem is to select a subset from a candidate set to optimize a given objective~\cite{qian2020distributed}. Based on different objectives, different subset set selection algorithms have been designed in the literature. For example, distance-based subset selection (DSS)~\cite{singh2019distance} is to maximize the uniformity level of the selected subset~\cite{shang2021distance}. Hypervolume subset selection (HSS)~\cite{bringmann2014two, kuhn2016hypervolume} and greedy HSS (GHSS)~\cite{bradst2007Incrementally, guerreiro2016greedy,bradst2006Maximising} are to maximize the hypervolume of the selected subset. In some studies~\cite{bringmann2014generic, ishibuchi2020new, ishibuchi2016compare}, subset selection is used to select a solution subset from an external archive as the final output to be presented to the decision maker. \textcolor{black}{These studies showed that solution sets which are better than the final population can be obtained by subset selection. Subset selection is also useful in EMO algorithms since we can choose an arbitrary number of solutions independent of the population size.} In our experiments, we use an efficient GHSS algorithm~\cite{chen2021fast} to find a final solution set from an external archive. 

\subsection{EMO Algorithm Framework with an External Archive}
\begin{figure}[b]
    \centering
    \includegraphics[width=1.0\linewidth]{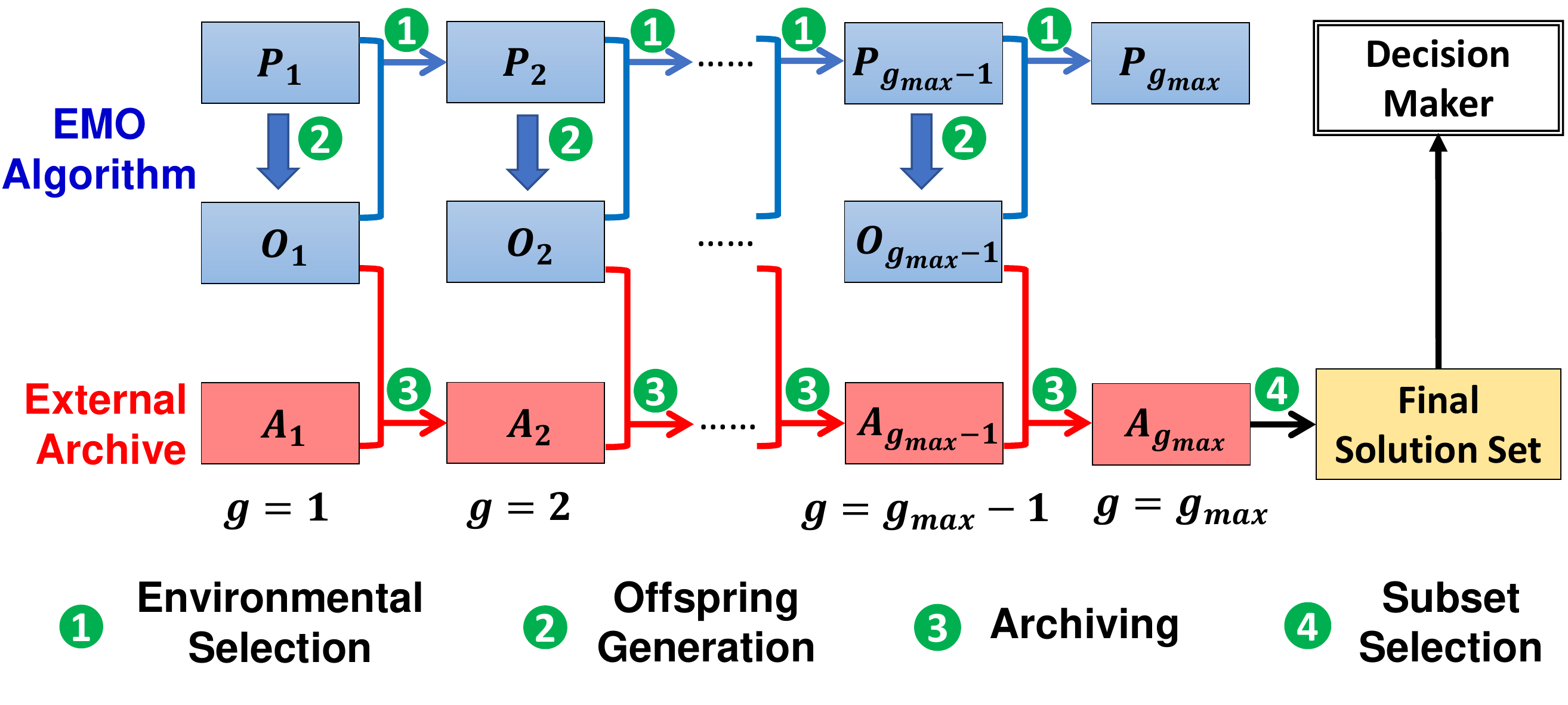}
    \caption{Our EMO algorithm framework with an external archive.}
    \label{fig:framework}
\end{figure}

Fig.~\ref{fig:framework} shows the EMO algorithm framework with an external archive. The blue part (i.e., the upper two blocks with the population $P_g$ and its offspring $O_g$ in each generation $g$) is a base EMO algorithm. The maximum number of generations, which is the termination condition, is denoted by $g_{max}$ in Fig.~\ref{fig:framework}. In each generation $g$ $(1\leq g \leq g_{max}-1)$, the offspring $O_g$ is generated from the population $P_g$. The next population $P_{g+1}$ is chosen from $P_g \cup O_g$ by an environmental selection mechanism. In most EMO algorithms, the final population $P_{g_{max}}$ is presented to the decision maker. The red part in Fig.~\ref{fig:framework} (i.e., the bottom block with the external archive $A_g$ in each generation $g$) is added to the base EMO algorithm to store examined solutions. The initial archive $A_1$ is the same as the initial population $P_1$ in the first generation. In each generation $g$ $(2\leq g \leq g_{max})$, the archive $A_g$ is obtained from $A_{g-1} \cup O_{g-1}$ by an archive maintenance mechanism. A final solution set, which is presented to the decision maker, is selected from the final archive $A_{g_{max}}$ by a subset selection mechanism.
Since the external archive has no influence on the base EMO algorithm (i.e., $A_g$ has no influence on $P_g$ and $O_g$ in Fig.~\ref{fig:framework}), it can be incorporated into almost all EMO algorithms. When a base EMO algorithm has an archive as in Table.~\ref{tab:archiveSizeTable}, both the current population and the archive of such an EMO algorithm can be viewed as the population $P_g$ in Fig.~\ref{fig:framework}. In this manner, the external archive $A_g$ can be incorporated to store examined solutions with no influence on the search behavior of the base EMO algorithm.\par
In Fig.~\ref{fig:framework}, we can use various archiving mechanisms. The simplest one is to add all offspring to the archive (if the archive size is unbounded). \textcolor{black}{Other state-of-the-art archiving mechanisms such as ArchiveUpdateHD proposed in ~\cite{hernandez2022bounded} for Hausdorff approximation can also be applied.} However, in many studies~\cite{zitzler1999multiobjective, Pradi2006anew}, only nondominated solutions are stored in the archive (i.e., all dominated solutions are removed), and the archive size is bounded (i.e., a truncation operation is used). In this paper, this archiving mechanism is referred to as the standard strategy where only nondominated solutions are stored in a bounded archive as shown in Algorithm~\ref{algo:standard}. In each generation $g$ $(2\leq g \leq g_{max})$, the archive $A_g$ is obtained from $A_{g-1} \cup O_{g-1}$. First, all dominated solutions are removed. If the number of remaining nondominated solutions is larger than the archive size $s$, a truncation operation is applied to remove excess nondominated solutions. 
\begin{algorithm}
\caption{Standard Archiving Strategy}
\label{algo:standard}
\begin{algorithmic}
\Require $A_{g-1}$ and $O_{g-1}$ (archive and offspring in generation $g-1$, respectively), $g$ (generation index), $s$ (archive size)
\Ensure $A_{g}$ (archive in generation g)

\State $A_{g} = A_{g-1} \cup O_{g-1}$
\State $A_{g} = RemoveD(A_{g})$ /*Remove dominated solutions*/ 
\If{$|A_{g}|>s$}
    \State $A_{g}=Truncate(A_{g},s)$ /*Select $s$ solutions using a truncation operation*/
\EndIf
\end{algorithmic}
\end{algorithm}



\section{Two New Archiving Strategies}\label{sec:Strategy}
 It is likely that a better final solution set can be obtained from a larger archive. Thus, from the viewpoint of the final solution set quality, a larger archive is beneficial. However, the increase in the archive size severely increases the computation time for the archive maintenance as we show later in Section~\ref{sec:Discussion}. To decrease the computation time, we propose two strategies called the lazy periodical strategy and the last $X$-generation strategy.
\subsection{Lazy Periodical Strategy}
In the standard archiving strategy in Algorithm~\ref{algo:standard}, dominated solutions are removed at every generation. If the archive is not full, it is not needed to remove dominated solutions unless the archive is used for other purposes such as the mating selection (e.g., Two\_Arch~\cite{Pradi2006anew}) and the weight vector adaptation (e.g., AdaW~\cite{Li2020what}). Therefore, if we remove dominated solutions from the archive only when the archive is full, the total computation time will decrease. We call this strategy the lazy strategy in this paper. Since the difference between the standard strategy and the lazy strategy is only the timing of the removal of the dominated solutions when the archive is not full, there is no difference in the handling of nondominated solutions between these two strategies. Thus, these strategies always store exactly the same nondominated solutions at each generation. As a result, we can obtain the same final solution set from them. For the same reason, they need the same computation time for both the truncation of the nondominated solutions and the selection of the final solution set. That is, the same final solution set is obtained by the lazy strategy with less computation time as the standard strategy.\par
Once the archive becomes full with nondominated solutions, there is no difference between these two strategies since the removal of the dominated solutions is needed at every generation. One idea for further decreasing the computation time is to perform the archive maintenance less frequently, i.e., at every $T$ generations instead of every generation. This idea is called the lazy periodical strategy in this paper, which is explained in Algorithm~\ref{algo:lazy_periodical}. In the lazy periodical strategy, the archive maintenance is performed at every $T$ generations. Thus, at every $T$ generations, the number of solutions in the achieve decreases from $s+NT$ to $s$ where $N$ is the population size and $s$ is the archive size. This means that the lazy periodical strategy needs the memory for storing $s + NT$ solutions whereas the standard strategy and the lazy strategy need the memory for $s + N$ solutions. In the final generation, the archive maintenance is always performed to decrease the number of solutions in the archive to the archive size $s$. The lazy strategy is a special case of the lazy periodical strategy with $T = 1$ (i.e., the archive maintenance is performed at every generation when the number of solutions in the archive is larger than the archive size $s$).
\begin{algorithm}
\caption{Lazy Periodical Strategy}
\label{algo:lazy_periodical}
\begin{algorithmic}
\Require $A_{g-1}$ and $O_{g-1}$ (archive and offspring in generation $g-1$, respectively), $g$ (generation index), $g_{max}$ (maximum number of generations), $s$ (archive size), $T$ (generation update interval)
\Ensure $A_{g}$ (archive in generation g)
\State $A_{g} = A_{g-1} \cup O_{g-1}$
\If{$(g_{max}-g) \mod T =0 $}
\If{$|A_{g}|>s$ or $g=g_{max}$}
\State$A_g$\,$=$\,$RemoveD(A_{g})$ /*Remove dominated solutions*/ 
\EndIf
\If{$|A_{g}|>s$}
\State $A_{g} = Truncate(A_{g},s)$ /*Select $s$ solutions using a truncation operation*/
\EndIf
\EndIf
\end{algorithmic}
\end{algorithm}
\vspace{-1.5em}
\subsection{Last $X$-Generation Strategy}
Solutions in later generations are likely to be better than those in earlier generations~\cite{nan2020reverse}. Based on this intuition, we propose a strategy called the last $X$-generation strategy, which simply stores all solutions in the last $X$ generations. In this strategy, the archive is initialized as the population $P_{g_{max}-X+1}$, and stores all offspring in the subsequent generations (from $O_{g_{max}-X+1}$ to $O_{g_{max}-1}$). Dominated solutions are removed from the archive in the final generation. When $X=1$, only the final population is stored in the archive. In this extreme case, the archive is the same as the final population. When $X=g_{max}$, all examined solutions are stored in the archive, and dominated solutions are removed in the final generation. This extreme case is the same as the lazy strategy with an unbounded archive where dominated solutions are removed only in the last generation (and the archive truncation is not used). In the last $X$-generation strategy, the value of $X$ is specified using the archive size $s$ as $X=\min\{\lfloor \frac{s}{N} \rfloor,g_{max}\}$ where $\lfloor  x \rfloor$ is the largest integer smaller than or equal to $x$. Thus, the total number of stored solutions in the archive is the same as or smaller than the archive size $s$. As a result, $s$ or less nondominated solutions are obtained from the last $X$-generation strategy. The specification of $X$ by $X=\min\{\lfloor \frac{s}{N} \rfloor,g_{max}\}$ is to remove the need for the archive truncation. We can also use a larger value than this specification (e.g., $X=20$ for $s=500$ and $N=100$). In this case, the lazy or lazy periodical strategy is needed in later generation in the last $X$ generations.
\subsection{Illustrations of Four Archiving Strategies}
Here we summarize the four archiving strategies.\par
\textbf{Standard Strategy}: Dominated solutions are removed at every generation. The truncation is performed if needed (i.e., if the number of remaining nondominated solutions in the archive exceeds the archive size).\par
\textbf{Lazy Strategy}: Dominated solutions are removed only when the number of solutions in the archive exceeds the archive size. Then, the truncation is performed if needed.\par
\textbf{Lazy Periodical Strategy}: The number of solutions in the archive is monitored at every $T$ generations. If it exceeds the archive size, dominated solutions are removed. Then, the truncation is performed if needed. Independent of the value of $T$, in the final generation, dominated solutions are removed, and the truncation is performed if needed.\par
\textbf{Last $\bm{X}$-Generation Strategy}: All solutions in the last $X$ generations are stored in the archive. Then, dominated solutions are removed in the final generation.\par
\begin{figure}[!h]
    \centering
    \subfigure[]{\includegraphics[width=0.4937\linewidth,trim=6 0 0 0,clip]{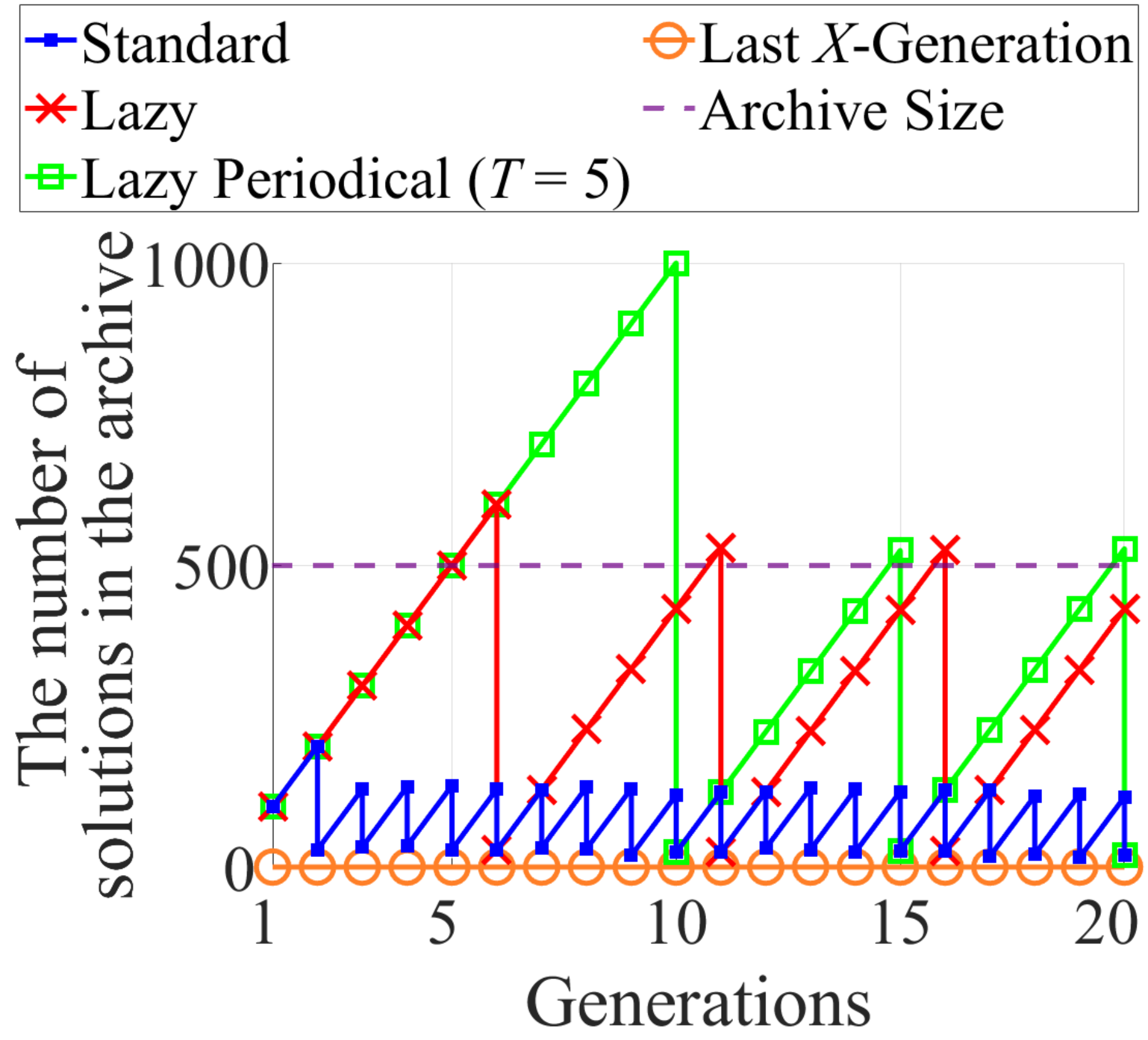}}
    \subfigure[]{\includegraphics[width=0.4937\linewidth,trim=6 0 0 0,clip]{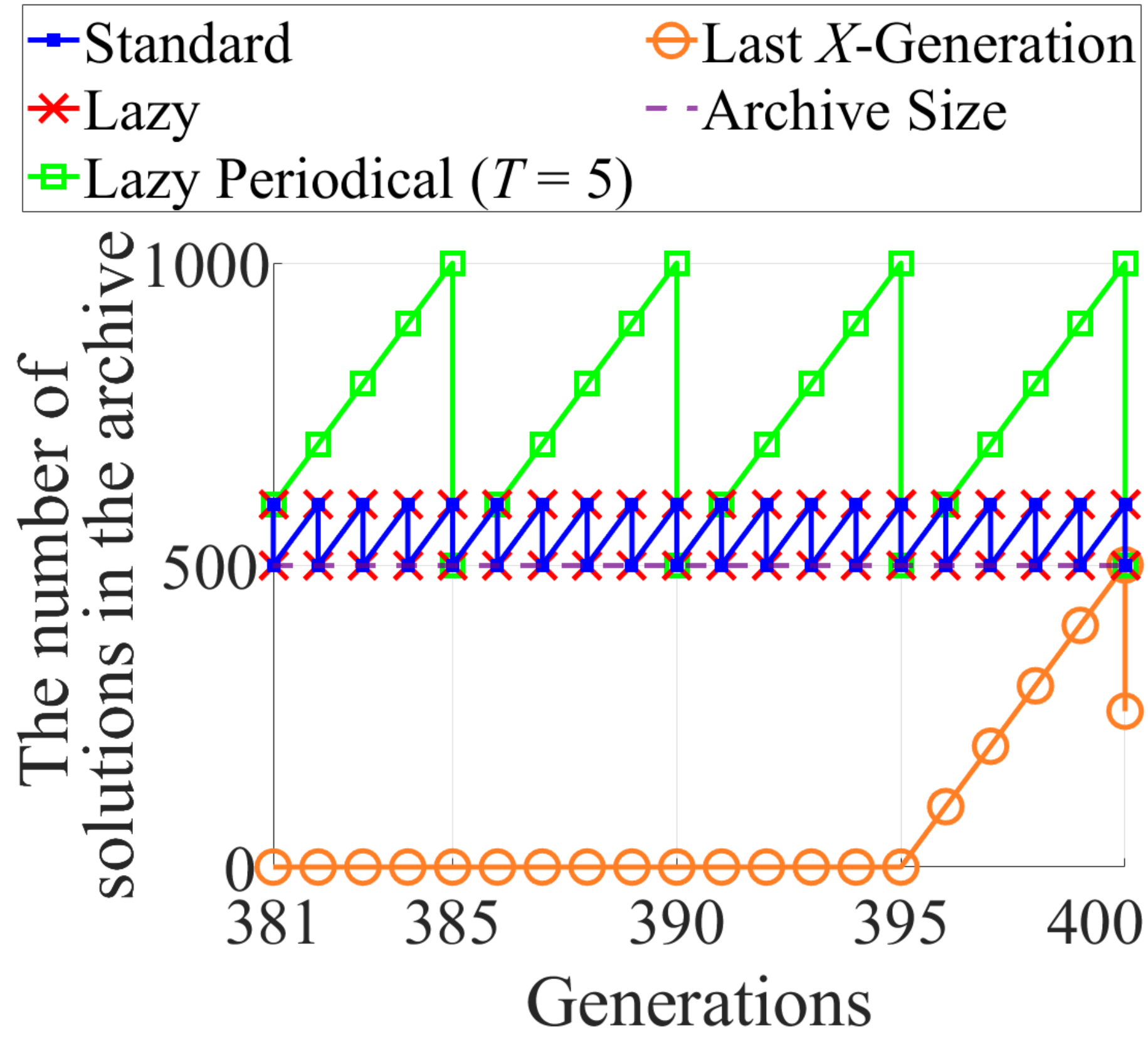}}
    \caption{The number of solutions in the archive by each archiving strategy. NSGA-II with the population size 100 is applied to 3-objective DTLZ1 for 400 generations. (a) In the first 20 generations. (b) In the last 20 generations.}
    \label{fig:NumberCompare}
\end{figure}

To demonstrate the archiving behavior of each strategy, we apply NSGA-II~\cite{deb2002fast} with the population size 100 to 3-objective DTLZ1\cite{deb2005scalable} for 400 generations (i.e., $N=100$ and $g_{max}=400$). The archive size $s$ is specified as $500$ (i.e., $s=500=5N$). Each of the four strategies is used for the archive maintenance. In the lazy periodical strategy, the archive update interval $T$ is specified as $T=5$. In the last $X$-generation strategy, $X$ is specified as $X = 5$ since$\lfloor \frac{s}{N} \rfloor=5$. Fig.~\ref{fig:NumberCompare} shows the number of solutions in the archive at each generation for each strategy. The sharp decrease of each curve in Fig.~\ref{fig:NumberCompare} means that the corresponding strategy maintains the archive by removing some solutions.
\par
If we use the standard strategy (blue line), the number of solutions in the archive is small in the first 20 generations in Fig.~\ref{fig:NumberCompare} (a) since dominated solutions are removed at every generation. The lazy strategy (red line) removes them only when the number of solutions in the archive exceeds the archive size 500 (purple dashed line). Whereas there exists a large difference between these two strategies in Fig.~\ref{fig:NumberCompare} (a), they have exactly the same archive after the removal of the dominated solutions. When the archive is full with nondominated solutions as in the last 20 generations in Fig.~\ref{fig:NumberCompare} (b), these two strategies show exactly the same behavior since the archive maintenance is performed at every generation even in the lazy strategy.\par
In the lazy periodical strategy with $T = 5$ (green line), the number of solutions in the archive drops at every 5 generations when it exceeds the archive size 500 (purple dashed line) in Fig.~\ref{fig:NumberCompare}. In the last $X$-generation strategy (orange line), the archive remains empty until the 395th generation. Then, it starts to store all examined solutions. The number of solutions in the archive drops in the final generation since dominated solutions are removed from the archive.\par
In Fig.~\ref{fig:NumberCompare}, the memory requirement in each strategy is shown by the peak value of the corresponding line. Each strategy needs the memory for storing $s+N$ solutions (the standard and lazy strategies), $s+TN$ solutions (the lazy periodical strategy), and $XN$ solutions (the last $X$-generation strategy) where $XN$ is the same as or smaller than $s$ since $X=min\{\lfloor \frac{s}{N} \rfloor,g_{max}\}$. When $T$ is not large and the archive size $s$ is much larger than the population size $N$, all strategies need similar memory size. This means that the memory requirement mainly depends on the archive size $s$ (not the choice of an archiving strategy). For example, if $N = 100$, $s = 5,000$ and $T = 5$, these four strategies need the memory size for storing 5,000-5,500 solutions.\par

\textcolor{black}{
In our experiments, an efficient tree-based nondominated sorting method (i.e., T-ENS~\cite{zhang2018decision}) is used to remove dominated solutions in all strategies. Details of T-ENS are shown in the supplementary file (Pages 30 and 31). When most solutions in the archive are nondominated, T-ENS has a time complexity of $O(M|A|\log |A|/\log M)$~\cite{zhang2018decision} where $|A|$ is the number of solutions in the archive $A$ and $M$ is the number of objectives. However, T-ENS is not the best method with respect to the time complexity. Alternatively, many other data structures (e.g., linear lists, ND-Tree~\cite{jaszkiewicz2018ND} and Dominance Decision Trees~\cite{schutze2003newdata}) can be used to remove dominated solutions in the archive as explained in~\cite{fieldsend2020data}. We examine a simple linear list structure, and its results are included in the supplementary file (Pages 30 and 31).}
\textcolor{black}{When the truncation is needed, we use a greedy distance-based inclusion algorithm~\cite{singh2019distance} in all strategies. Its time complexity is quadratic with respect to the archive size. We also examine a greedy crowding distance-based removal algorithm for the truncation, and its details and results are included in the supplementary file (Pages 32 and 33). For the selection of the final solution set, a lazy greedy inclusion hypervolume subset selection (LGI-HSS) algorithm~\cite{chen2021fast} is used in all strategies. The reference point for hypervolume calculation in LGI-HSS is specified as $(1.2, 1.2, ..., 1.2)$ in the normalized objective space where the estimated ideal and nadir points from the final archive are $(0, 0, ..., 0)$ and $(1, 1, ..., 1)$, respectively. The final solution set size is the same as the population size in our experiments.}

\section{Settings of Computational Experiments}\label{sec:Experiment}
We examine the effects of the archive size in the four archiving strategies on the quality of the selected solution set and the total computation time through computational experiments. Their settings are explained in this section.
\subsection{EMO Algorithms and Test Problems}\label{sec:Test data}
We use three EMO algorithms: NSGA-II~\cite{deb2002fast}, MOEA/D-PBI~\cite{zhang2007moea} and NSGA-III~\cite{deb2014evolutionary}. For test problems, we choose DTLZ1-4~\cite{deb2005scalable} and their minus versions (Minus-DTLZ1-4)~\cite{ishibuchi2017performance} with 3, 5 and 8 objectives ($M=3,5,8$), i.e., 24 instances in total. The population size $N$ is set as 91, 210, and 156 for the test problems with 3, 5, and 8 objectives, respectively. The termination condition $g_{max}$ for each test problem is summarized in Table~\ref{tab:generation_settings}. Each algorithm is applied to each test problem 21 times (i.e., 21 independent runs). In each run, the current and offspring populations in each generation are stored. As a result, we have 1,512 sequences of the current and offspring populations (i.e., 3 algorithms $\times$ 24 test problems $\times$ 21 runs $=$ 1,512 sequences). The length of each sequence is the same as the maximum number of generations in the corresponding run. These sequences are used to examine the four archiving strategies under various specifications of the archive size $s$ and the archive update interval $T$. Since our external archive has no effect on the search behavior of each EMO algorithm, we can examine all four strategies using the same 1,512 sequences. The average results over 21 runs (e.g., 21 sequences) are calculated for each of the 72 combinations of the 3 EMO algorithms and the 24 test problems. Thus, we have 72 average results for each strategy with various specifications. 
\begin{table}[!h]
    \centering
        \caption{Maximum number of generations $G_{max}$ (Termination condition)}
    \begin{tabular}{|c|c|c|c|}
    \hline
        \multirow{2}{*}{Problem} & \multicolumn{3}{c|}{Number of objectives: $M$}\\ \cline{2-4}
         & 3 & 5  & 8 \\ \hline
        DTLZ1 &400 &600&750\\ \hline
        DTLZ2 &250 &350&500\\ \hline
        DTLZ3 &1,000 & 1,000&1,000\\ \hline
        DTLZ4 &600 &1,000 &1,250\\ \hline
        Minus-DTLZ1 &400 &600&750\\ \hline
        Minus-DTLZ2 &250 &350&500 \\ \hline
        Minus-DTLZ3 &1,000 & 1,000&1,000\\ \hline
        Minus-DTLZ4 &600 &1,000 &1,250\\ \hline
    \end{tabular}
    \label{tab:generation_settings}
\end{table}
\subsection{Archive Size and Archive Update Interval}
In the lazy periodical strategy, five specifications of the archive update interval $T$ are examined: $T=1$, $2$, $5$, $10$ and $20$. When $T=1$, the lazy periodical strategy is the same as the lazy strategy. In all four strategies, a wide variety of archive size specifications are examined: $s=N$, $2N$, $5N$, $10N$, $20N$, $50N$, $100N$, $200N$, $500N$, $1,000N$ and $2,000N$ where $N$ is the population size. When $s\geq g_{max}N$, all solutions can be stored in the archive. However, in the standard strategy, dominated solutions are removed at every generation. In all the other strategies, dominated solutions are removed only in the final generation when $s\geq g_{max}N$. 
\subsection{Performance Metric}
To evaluate the quality of the selected final solution set, we use the hypervolume indicator. The reference point is set as $(1.2, 1.2, ..., 1.2)$ in the normalized objective space where the true ideal and nadir points are $(0, 0, ..., 0)$ and $(1, 1, ..., 1)$, respectively. Note that the true ideal and nadir points are used in the normalization for the performance evaluation whereas they are estimated in each run using the final archive (i.e., candidate solution set) to select the final solution set.\par
To evaluate the efficiency of each archiving strategy, we record the total computation time for the three operations: the removal of dominated solutions, the truncation of nondominated solutions, and the selection of a final solution set. The available runtime (i.e., upper bound) is set as one hour. If the final solution set is not obtained within one hour, the execution of the archiving strategy is terminated. In this case, we have no results.\par
We perform all experiments on a virtual machine equipped with two ADM EPYC 7702 128-Core CUP@2.4GHz, 256GB RAM and Ubuntu Operating System. All codes are implemented in MATLAB R2021b and available from \url{https://github.com/HisaoLabSUSTC/ArchiveSize}. The implementation of the EMO algorithms is based on PlatEMO~\cite{tian2017platemo}.
\section{Experimental Results and Discussions}\label{sec:Discussion}
\subsection{Effects of the Archive Size on the Final Solution Set Quality}~\label{sec:Discussion_quality}
As explained in the previous section, the average hypervolume value of the selected final solution sets obtained by each archiving strategy with each archive size specification is calculated over 21 runs for each of the 72 combinations of the 3 algorithms and the 24 test problems. Experimental results on all 72 combinations are included in the supplementary file. Among them, Fig.~\ref{fig:hypervolume} shows nine results on three test problems (3-objective Minus-DTLZ1, 5-objective DTLZ3 and 8-objective Minus-DTLZ2). In the lazy periodical strategy, the archive update interval $T$ is specified as $T = 10$. Various specifications of the archive size $s$ are examined from the population size $N$ (the smallest archive size) to $2,000N$ (the largest archive size which is the same as the unbounded archive). As explained in Section~\ref{sec:Strategy}, the selected final solution sets are always the same between the standard strategy and the lazy strategy. Thus, their results are always the same in Fig.~\ref{fig:hypervolume}. For comparison, the average hypervolume value of the final population is also shown in each figure by a purple dashed line. In Fig.~\ref{fig:hypervolume}, some results are missing. For example, in Fig.~\ref{fig:hypervolume} (i), only the results obtained by the last $X$-generation and lazy periodical strategies are shown when $s = 100N$. No final solution sets are obtained by the other strategies within one hour. This issue is discussed in the next section.\par
\begin{figure*}[t!]
\centering
 \subfigure[NSGA-II on 3-objective Minus-DTLZ1] {\includegraphics[width=0.305\textwidth,trim=5 0 20 20,clip]{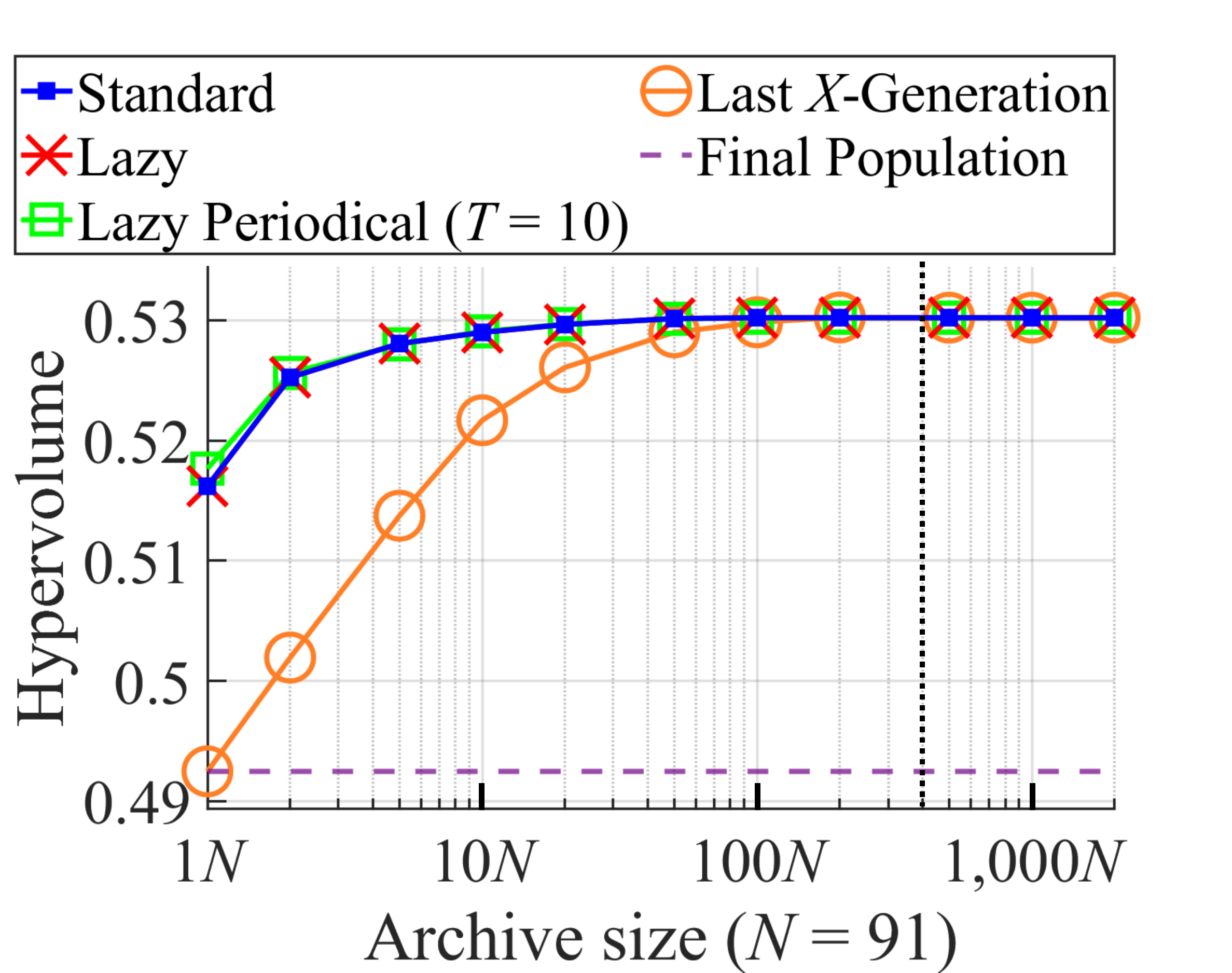}}
  \subfigure[MOEA/D-PBI on 3-objective Minus-DTLZ1 ] {\includegraphics[width=0.305\textwidth,trim=5 0 20 20,clip]{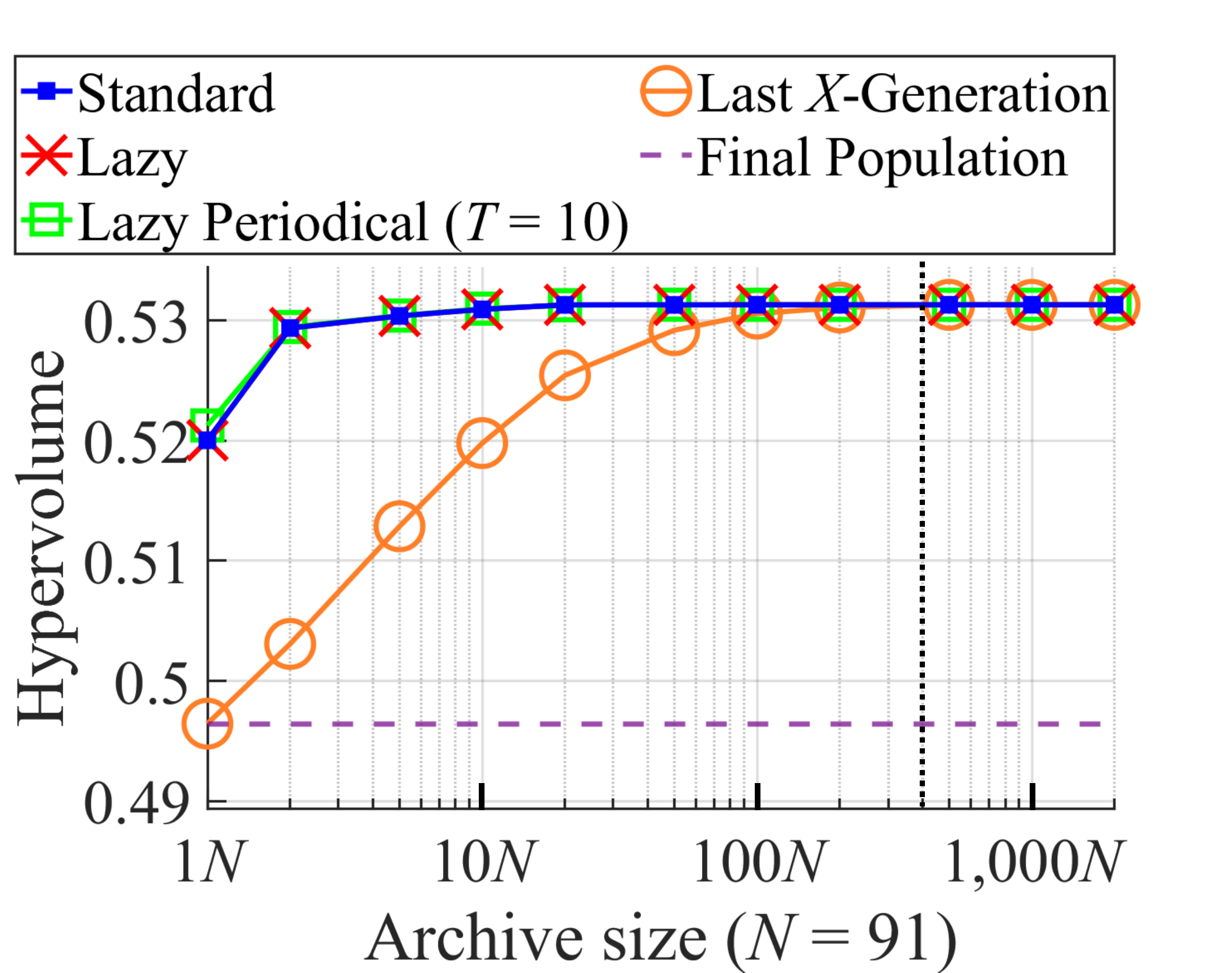}}
  \subfigure[NSGA-III on 3-objective Minus-DTLZ1 ]{\includegraphics[width=0.305\textwidth,trim=5 0 20 20,clip]{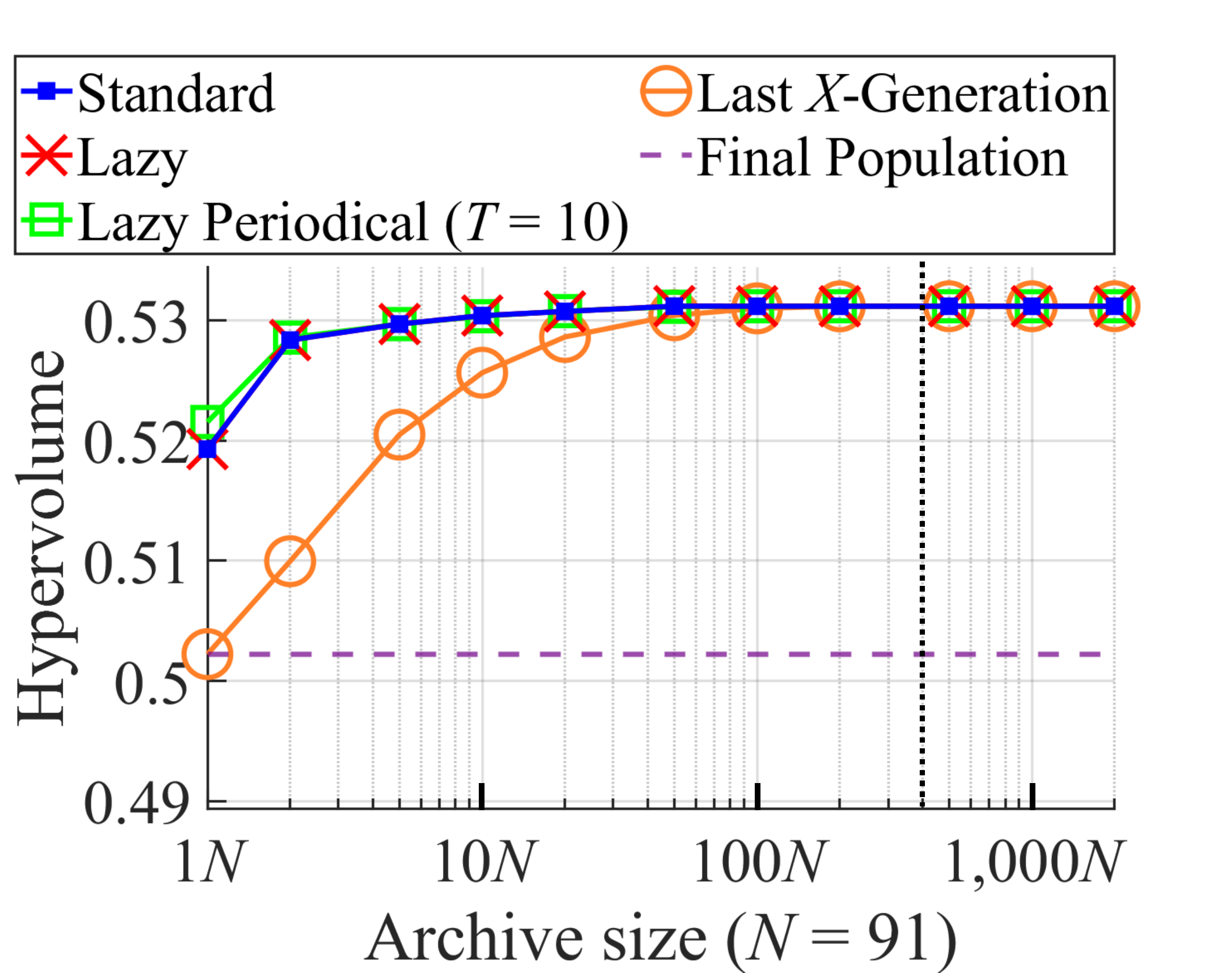}}\\
   \subfigure[NSGA-II on 5-objective DTLZ3 ]
  {\includegraphics[width=0.305\textwidth,trim=5 0 20 20,clip]{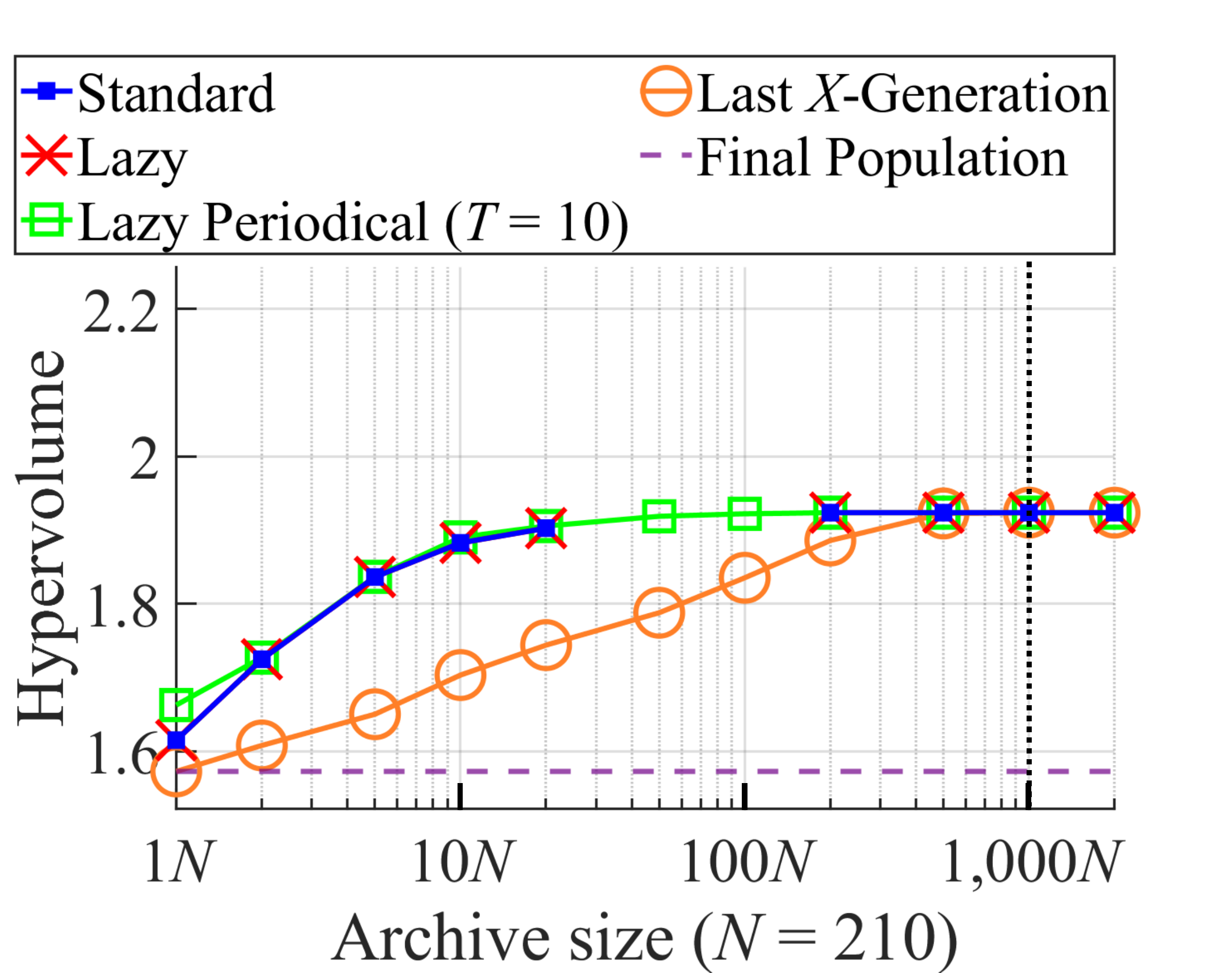}}
  \subfigure[MOEA/D-PBI on 5-objective DTLZ3] {\includegraphics[width=0.305\textwidth,trim=5 0 20 20,clip]{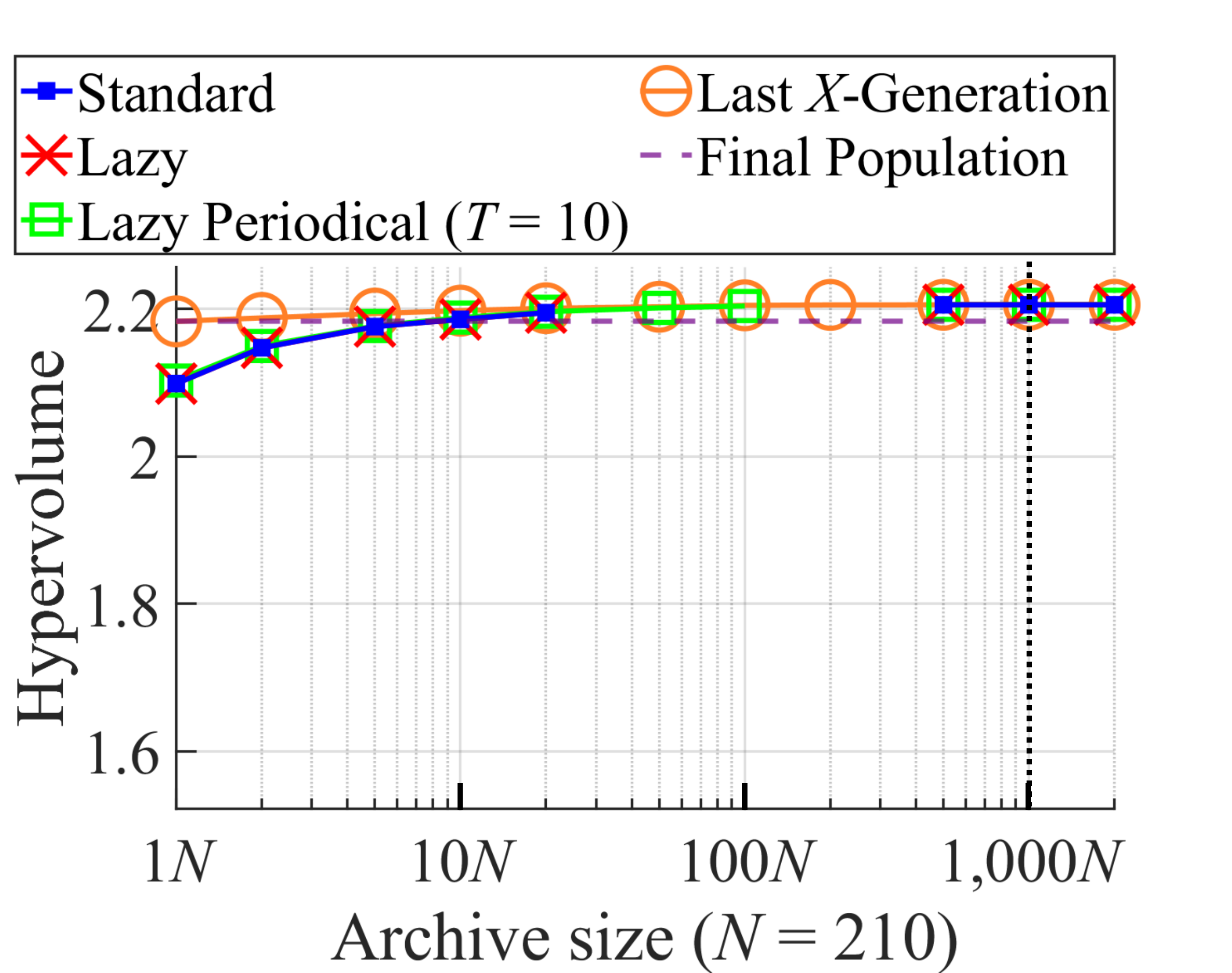}}
  \subfigure[NSGA-III on 5-objective DTLZ3 ]{\includegraphics[width=0.305\textwidth,trim=5 0 20 20,clip]{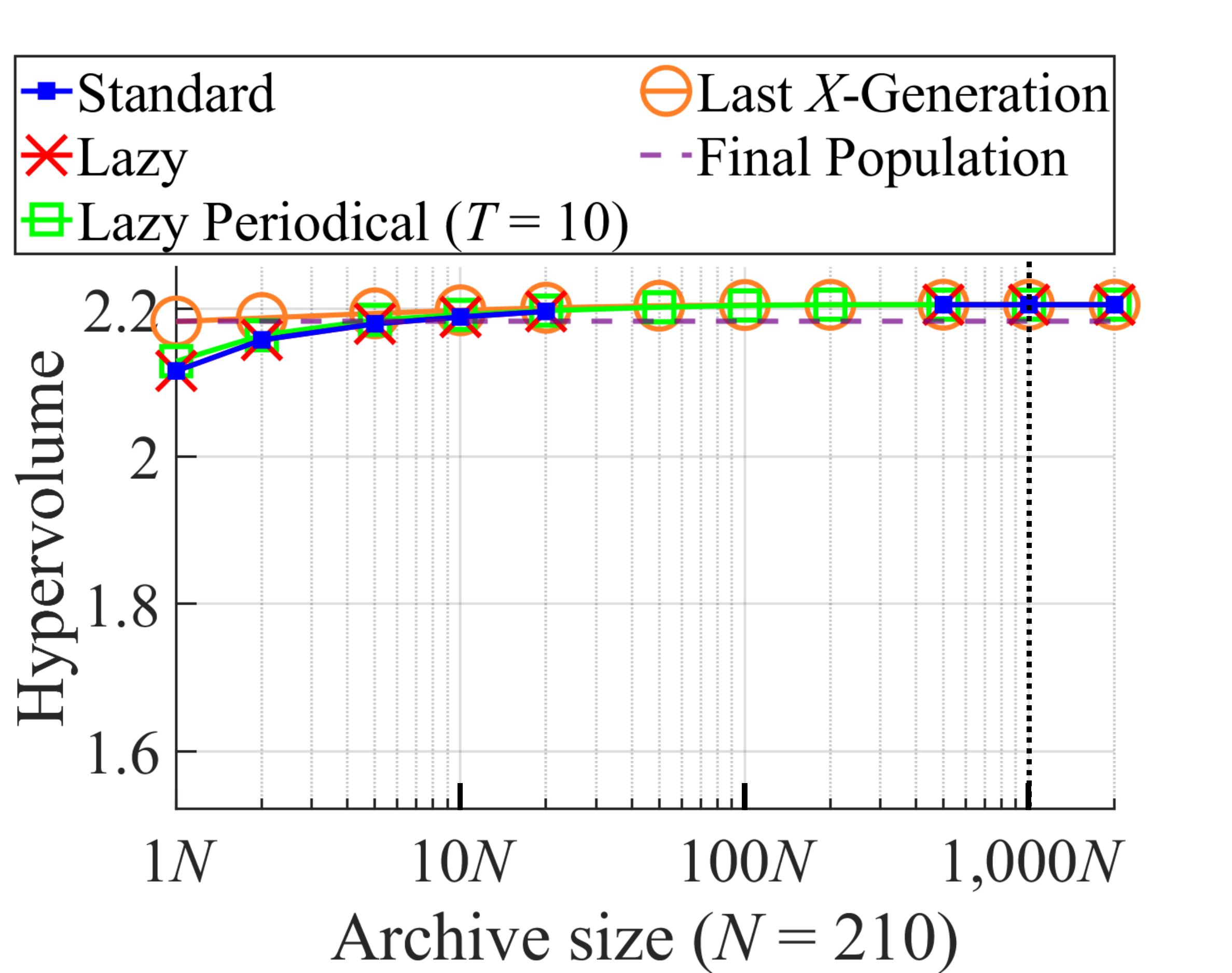}}\\
  \subfigure[NSGA-II on 8-objective Minus-DTLZ2 ]
  {\includegraphics[width=0.305\textwidth,trim=5 0 20 20,clip]{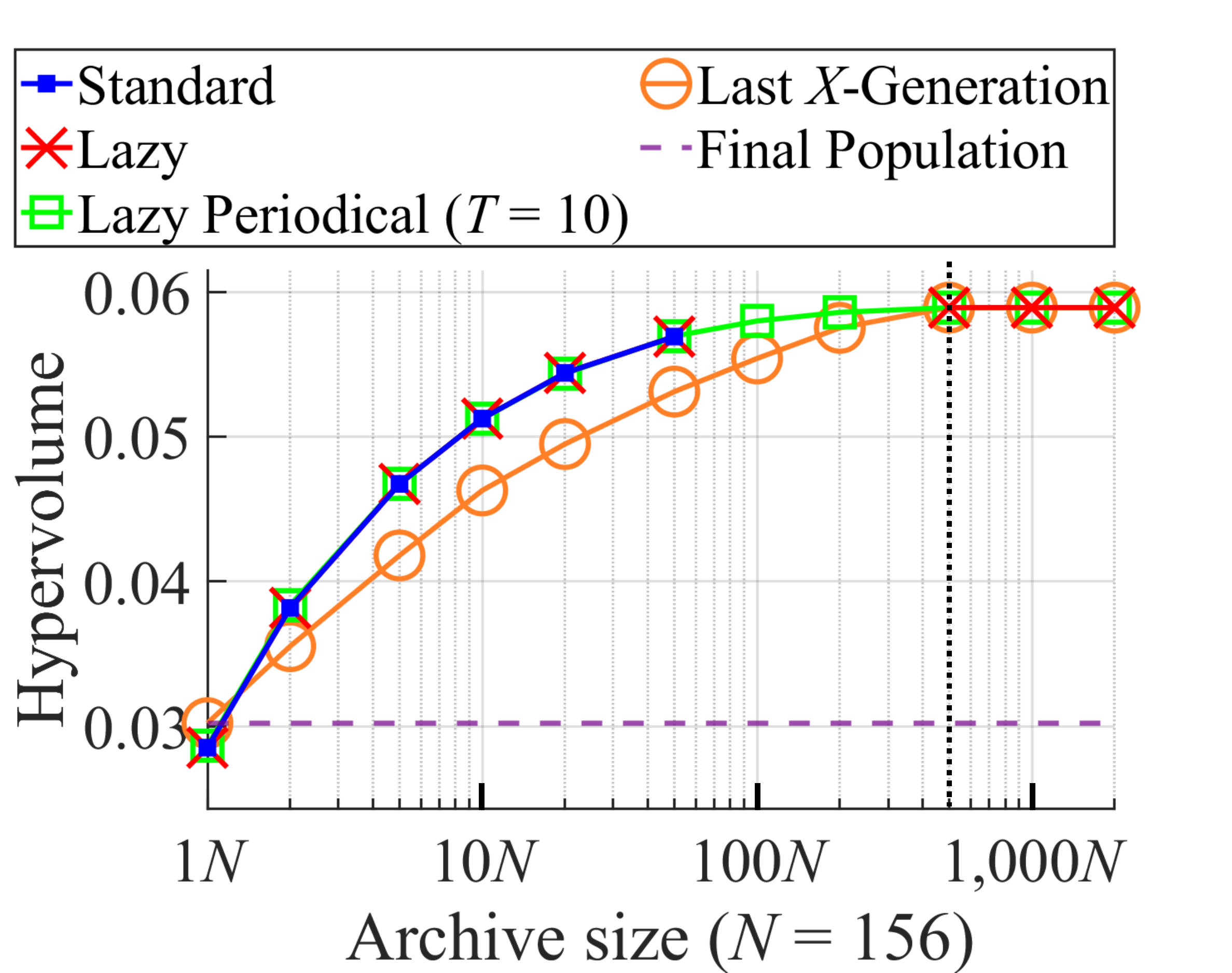}}
  \subfigure[MOEA/D-PBI on 8-objective Minus-DTLZ2] {\includegraphics[width=0.305\textwidth,trim=5 0 20 20,clip]{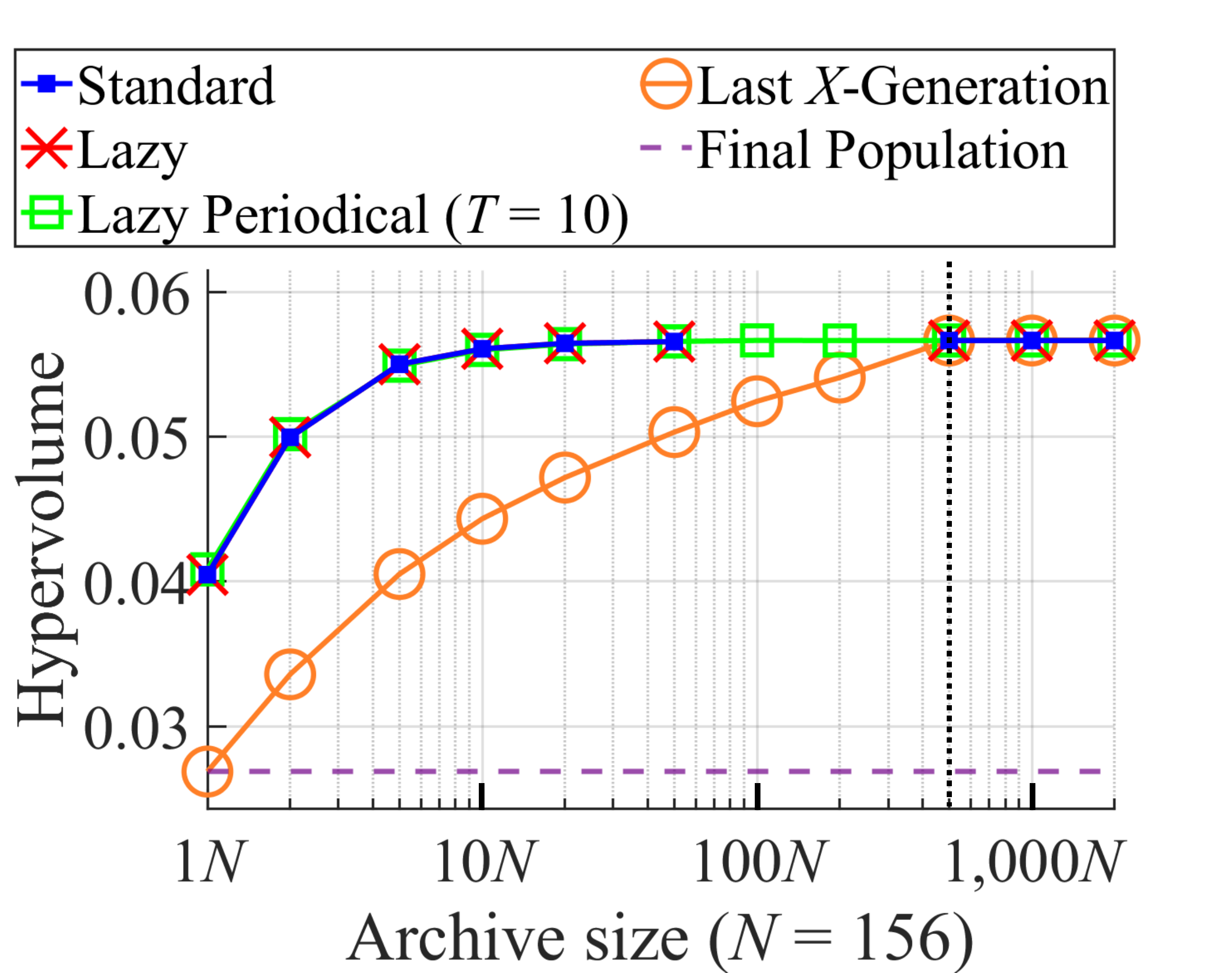}}
  \subfigure[NSGA-III on 8-objective Minus-DTLZ2 ]{\includegraphics[width=0.305\textwidth,trim=5 0 20 20,clip]{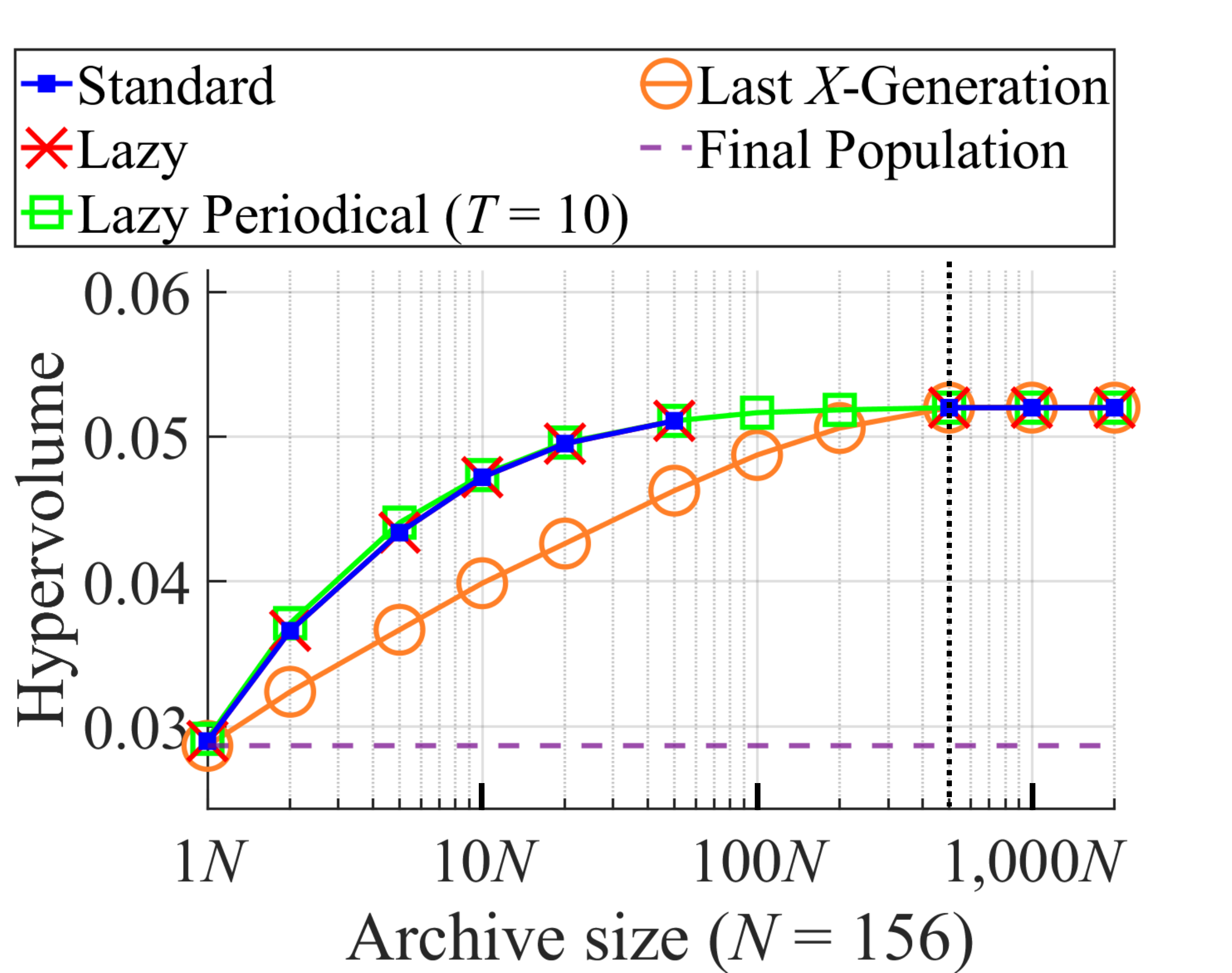}}
  \caption{Average hypervolume value of the selected final solution sets obtained by each archiving strategy with each archive size. \color{black}{The black vertical dotted line shows the total number of examined solutions in each run. When the archive size is larger than this line, the archive is equivalent to an unbounded archive.}}\label{fig:hypervolume}
  \vspace{-1em}
\end{figure*}
As shown by the top three figures (a)-(c) and the bottom three figures (g)-(i) in Fig.~\ref{fig:hypervolume}, in many cases, the quality of the selected final solution set is improved by increasing the archive size. The selected solution set is usually better than the final population. Since it is difficult for NSGA-II to find well-distributed solutions in a high-dimensional objective space with three or more objectives, its final population quality is clearly improved by subset selection in Fig.~\ref{fig:hypervolume}. For all Minus-DTLZ problems, MOEA/D-PBI and NSGA-III cannot find well-distributed solutions. As a result, the final population quality of these algorithms is also clearly improved by subset selection in Fig.~\ref{fig:hypervolume} for Minus-DTLZ1 and Minus-DTLZ2. On the contrary, the improvement of the final population quality is minor in Fig.~\ref{fig:hypervolume} (e) and (f). This is because MOEA/D-PBI and NSGA-III can find well-distributed solutions for DTLZ3. When the archive size is small in Fig.~\ref{fig:hypervolume} (e) and (f), the selected final solution sets obtained by the lazy and lazy periodical strategies are worse than the final population. This is because the environmental selection mechanisms of MOEA/D-PBI and NSGA-III work better than the greedy distance-based inclusion algorithm for DTLZ3. In other words, since the weight (reference) vector distribution in these algorithms is compatible with the shape of the Pareto front of DTLZ3, the obtained final populations are better than the selected final solution sets obtained from small archives maintained by the distance-based algorithm. However, even in Fig.~\ref{fig:hypervolume} (e) and (f), the final solution sets selected from large archives are better than the final populations.\par
With respect to the comparison among the four strategies, we can say that almost the same results are obtained from the four strategies when the archive size is very large (e.g., when it is larger than $500N$). Independent of the archive size, the quality of the selected final solution set of the lazy periodical strategy is similar to that of the standard and lazy strategies. The quality of the selected final solution set of the last $X$-generation strategy directly depends on the final population quality as shown in Fig.~\ref{fig:hypervolume} when the archive size is small (e.g., $s < 10N$).\par
\textcolor{black}{Some unusual behaviors are observed in our experimental results in the supplementary file. One is that the final population is better than the selected final solution sets even when the archive size is very large in some cases. Another is that the quality of the selected final solution set is deteriorated as the archive size increases in some cases. The details and explanations of the two unusual behaviors are included in the supplementary file.}\par
When no solutions are close to the Pareto front (e.g., 8-objective DTLZ3 with NSGA-II in page 6 of the supplementary file), the hypervolume value is always zero. However, in some other cases (e.g., 8-objective DTLZ4 with NSGA-II in page 6 of the supplementary file), even when the hypervolume value of the final population is zero, some selected final solution sets have positive hypervolume values. This is because the environmental selection mechanism of NSGA-II cannot store good solutions close to the Pareto front (due to the existence of dominance resistant solutions).

\subsection{Effects of the Archive Size on the Computation Time}\label{sec:tmp}
\begin{figure}[!b]
    \vspace{-1em}
 \centering
 \subfigure[3-objective Minus-DTLZ1] {\includegraphics[width=0.49\linewidth,trim=0 0 0 0,clip]{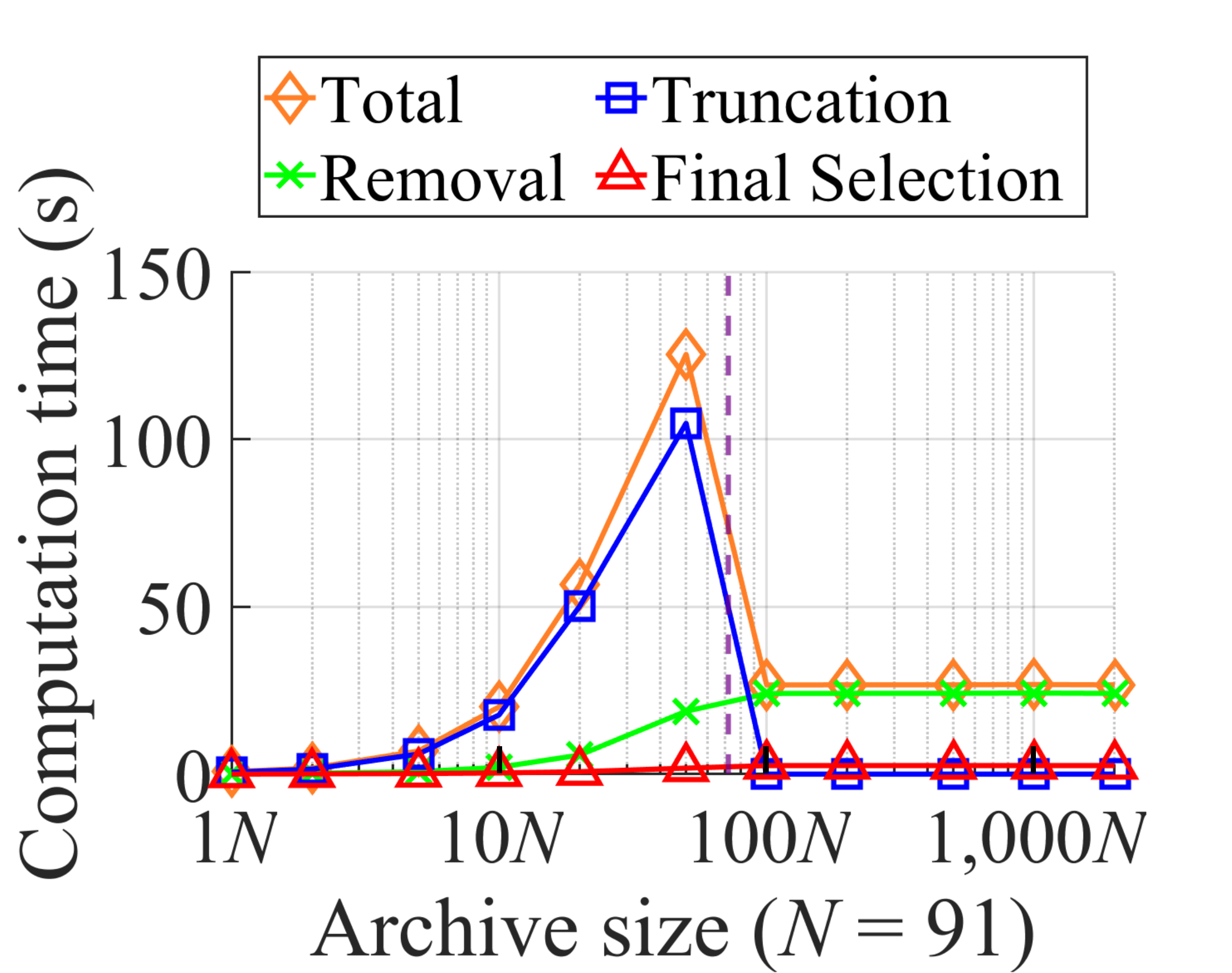}}
  \subfigure[5-objective DTLZ3]
  {\includegraphics[width=0.49\linewidth,trim=0 0 0 0,clip]{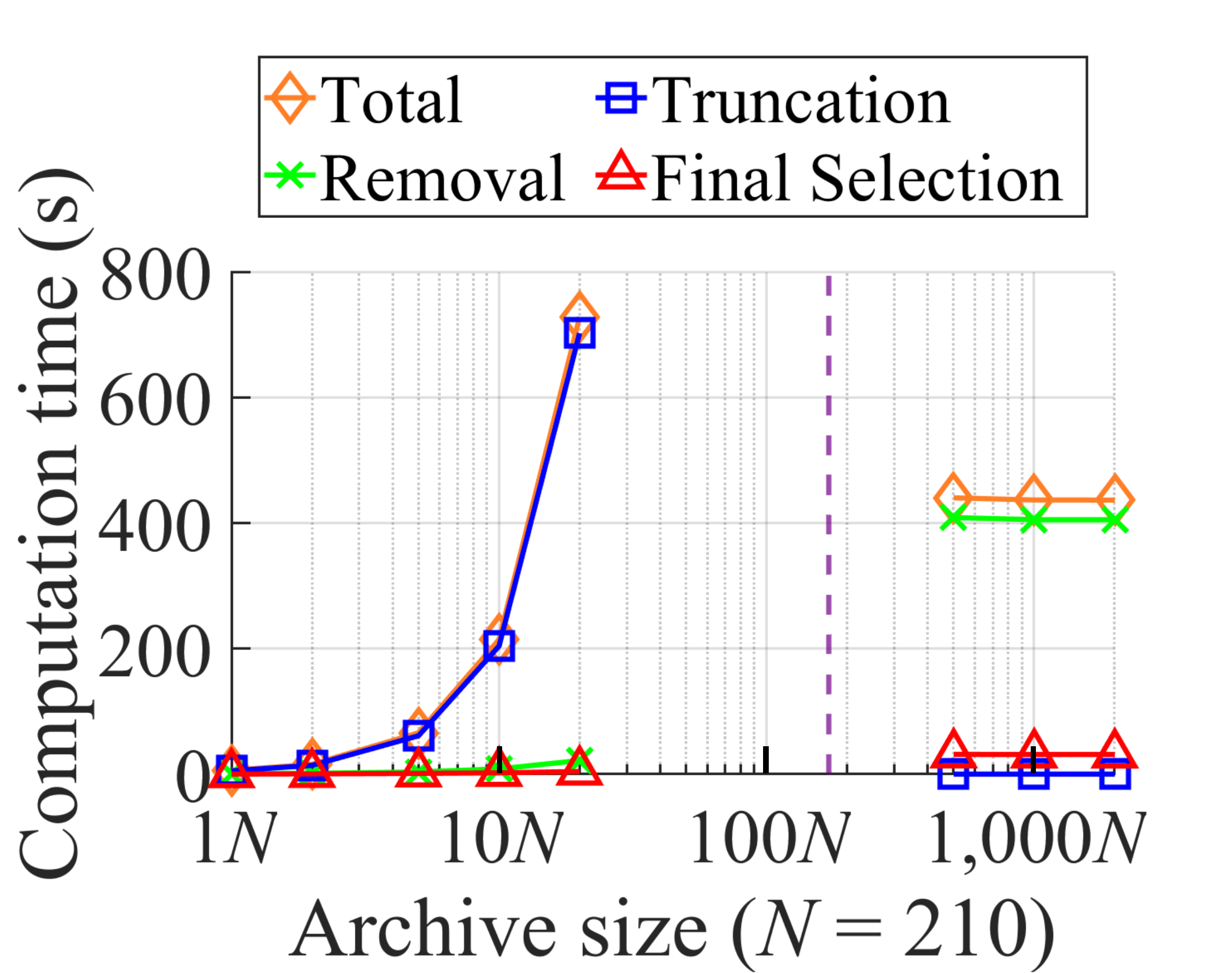}}
    \caption{Average computation time of the standard strategy for the 3-objective Minus-DTLZ1 and 5-objective DTLZ3 problems with NSGA-III. \color{black}{The purple vertical dashed line shows the average maximum number of nondominated solutions in the unbounded archive over 21 runs.}}
    \label{fig:standard_time}
 \end{figure}

As shown in Fig.~\ref{fig:hypervolume} (g)-(i), no final solution set is obtained within one hour when the standard and lazy strategies are used for large archives. In order to further examine the computation time, we calculate the average computation time for each of the three operations in the standard strategy: dominated solution removal, archive truncation, and final solution set selection. This is performed for all 72 combinations of the 3 EMO algorithms and the 24 test problems. Experimental results are included in the supplementary file. Among them, two experimental results are shown in Fig.~\ref{fig:standard_time} for the case of 3-objective Minus-DTLZ1 and 5-objective DTLZ3 with NSGA-III. The computation time for each operation and the total computation time are shown in Fig.~\ref{fig:standard_time} for various specifications of the archive size $s$.\par
In Fig.~\ref{fig:standard_time} (a) on 3-objective Minus-DTLZ1, the total computation time is large for the medium-size archive. When the archive size $s$ is $50N$, the total computation time is the largest. We can also see that most computation time is used by the truncation when $s = 50N$. By further increasing the archive size, the computation time for the truncation drastically decreases. This is because the number of nondominated solutions in the archive becomes smaller than the archive size. \textcolor{black}{In Fig.~\ref{fig:standard_time}, the maximum number of nondominated solutions in the unbounded archive is shown by a purple dashed vertical line.} Due to the large computation time for the truncation, our computational experiment cannot be completed within one hour when the archive size is specified between $50N$ and $200N$ in Fig.~\ref{fig:standard_time} (b). From Fig.~\ref{fig:standard_time}, we can see that the computation time severely increases by the increase in the archive size in the standard strategy. We can also see that the computation time is mainly used by the truncation (when the archive size is smaller than the total number of nondominated solutions) and the removal of the dominated solutions (when the archive size is larger than the total number of nondominated solutions).\par
\begin{figure}[!b]
      \vspace{-1em}
 \centering
 \subfigure[3-objective Minus-DTLZ1] {\includegraphics[width=0.49\linewidth,trim=0 0 40 0,clip]{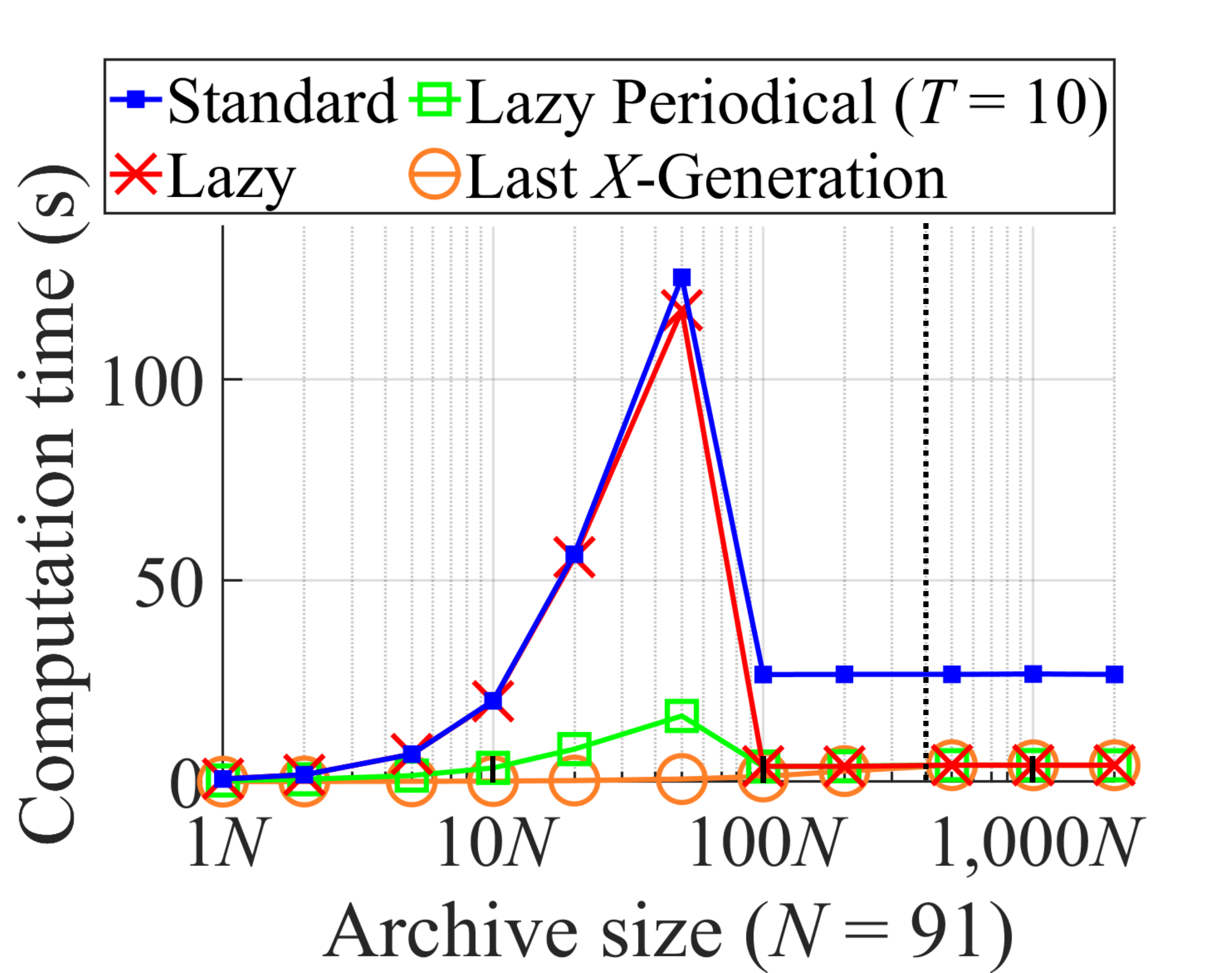}}
  \subfigure[5-objective DTLZ3] {\includegraphics[width=0.49\linewidth,trim=0 0 40 0,clip]{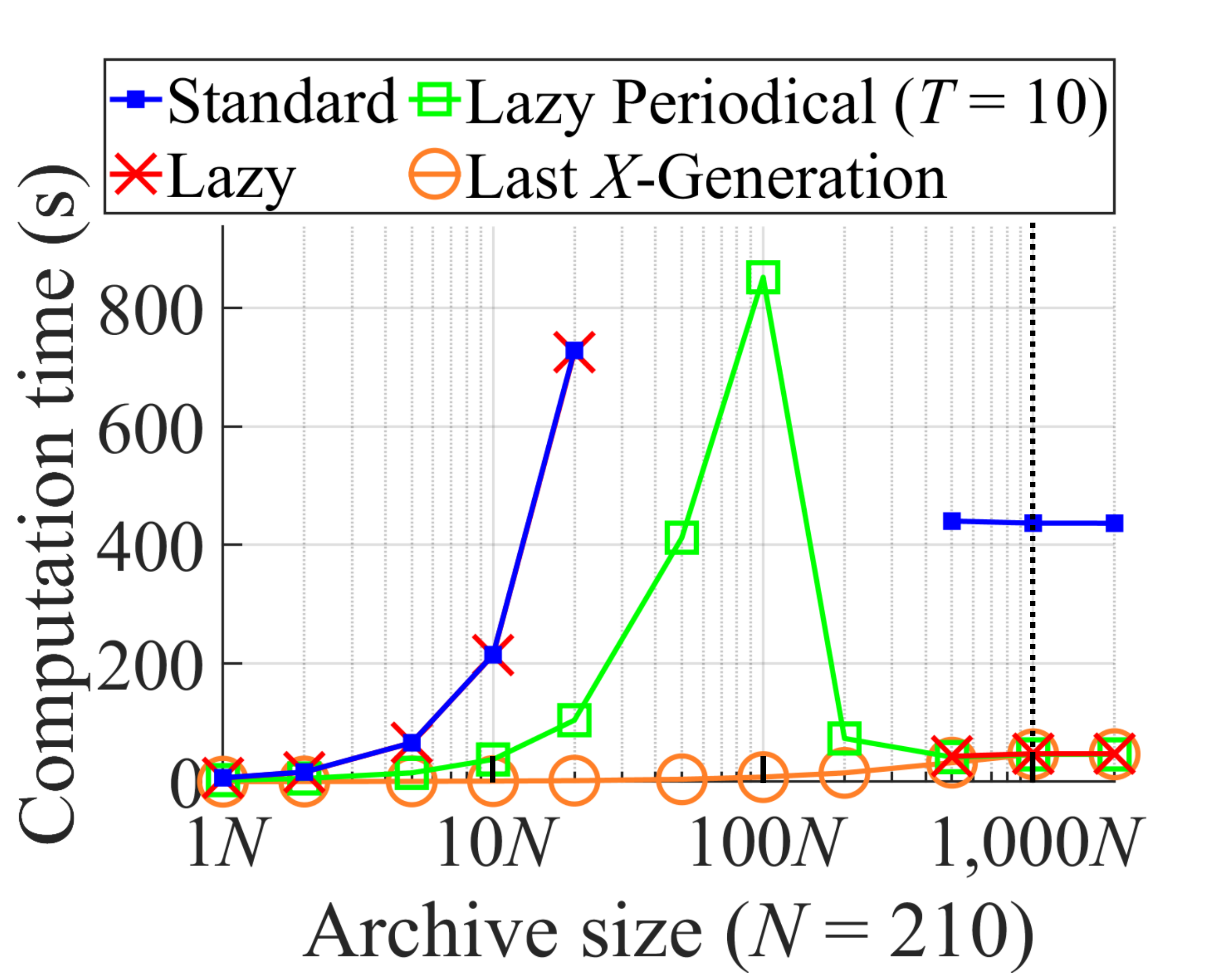}}
    \caption{Average computation time of each of the four strategies on the same two cases as in Fig.~\ref{fig:standard_time}. The total computation time is shown for each strategy. \color{black}{The black vertical dotted line shows the total number of examined solutions in each run. For comparison, the average computation time of NSGA-III is 1.2 seconds and 7.0 seconds in (a) and (b), respectively. }}\label{fig:time}

 \end{figure}

In Fig.~\ref{fig:time}, the total computation time for each of the four strategies is shown for the same two cases as in Fig.~\ref{fig:standard_time}. The archive update interval $T$ is specified as $T = 10$ in the lazy periodical strategy. Since the archive is updated at every 10 generations, the lazy periodical strategy is much faster than the standard and lazy strategy. In the lazy strategy, the archive is updated at every generation after the archive becomes full with nondominated solutions. Thus, its computation time is almost the same as that of the standard strategy when the archive size is small. The last $X$-generation strategy updates the archive only in the final generation. Thus, it is much faster than all the other strategies. When the archive size is larger than the total number of examined solutions (e.g., $s = 2,000N$ in Fig.~\ref{fig:time}), all strategies except for the standard strategy have the same computation time since they do not update the archive until the final population.
  \par
\begin{figure}[!t]
 \centering
 \subfigure[Average hypervolume value] {\includegraphics[width=0.49\linewidth,trim=0 0 40 0,clip]{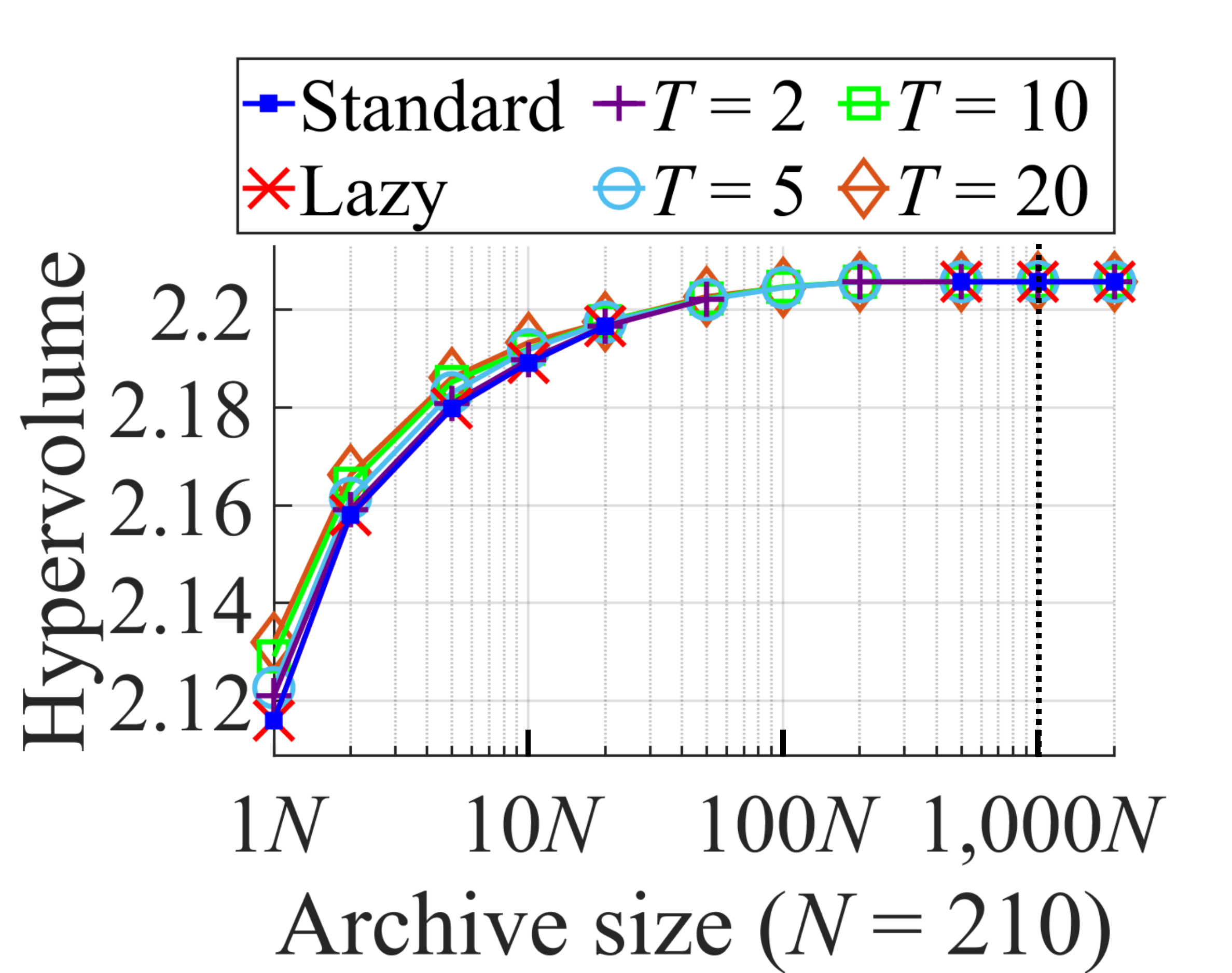}}
  \subfigure[Computation time] {\includegraphics[width=0.49\linewidth,trim=0 0 40 0,clip]{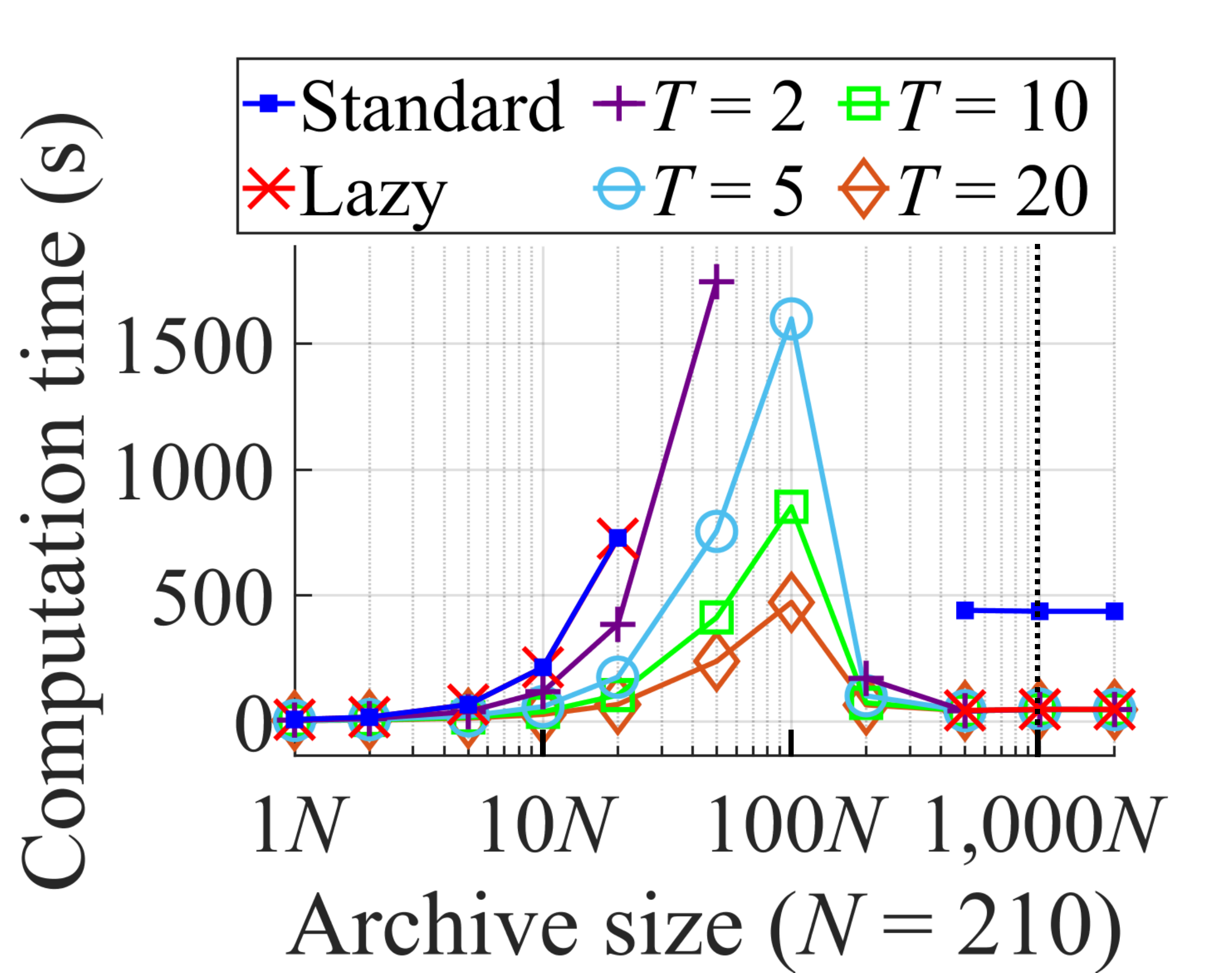}}
    \caption{Experimental results on 5-objective DTLZ3 with NSGA-III by the lazy periodical strategy with $T = 1, 2, 5, 10, 20$. For comparison, results obtained by the standard strategy are also shown. The lazy periodical strategy with $T = 1$ is the same as the lazy strategy. \color{black}{The black vertical dotted line shows the total number of examined solutions in each run.}}\label{fig:periodical}
    \vspace{-1em}
 \end{figure}

To further examine the effects of the archive update interval $T$ in the lazy periodical strategy, we show its experimental results on 5-objective DTLZ3 in Fig.~\ref{fig:periodical} where $T$ is specified as $T = 1, 2, 5, 10$ and $20$. For comparison, results obtained by the standard strategy are also shown. Fig.~\ref{fig:periodical} (a) shows that the final solution set quality is almost the same between the standard strategy and the lazy periodical strategy independent of the specification of $T$. This observation is consistent with the experimental results in Fig.~\ref{fig:hypervolume}. Fig.~\ref{fig:periodical} (b) shows that the increase of the archive update internal $T$ clearly decreases the computation time. When the archive size is larger than the total number of examined solutions (i.e., $s=2,000N$), the effect of $T$ on the computation time is almost zero. This is because the archive can store all nondominated solutions (i.e., no truncation is needed).
\subsection{Relation among Quality, Time, and Memory Size}
\begin{figure}[!h]
 \centering
 \subfigure[Quality and computation time.] {\includegraphics[width=0.49\linewidth,trim=0 0 0 0,clip]{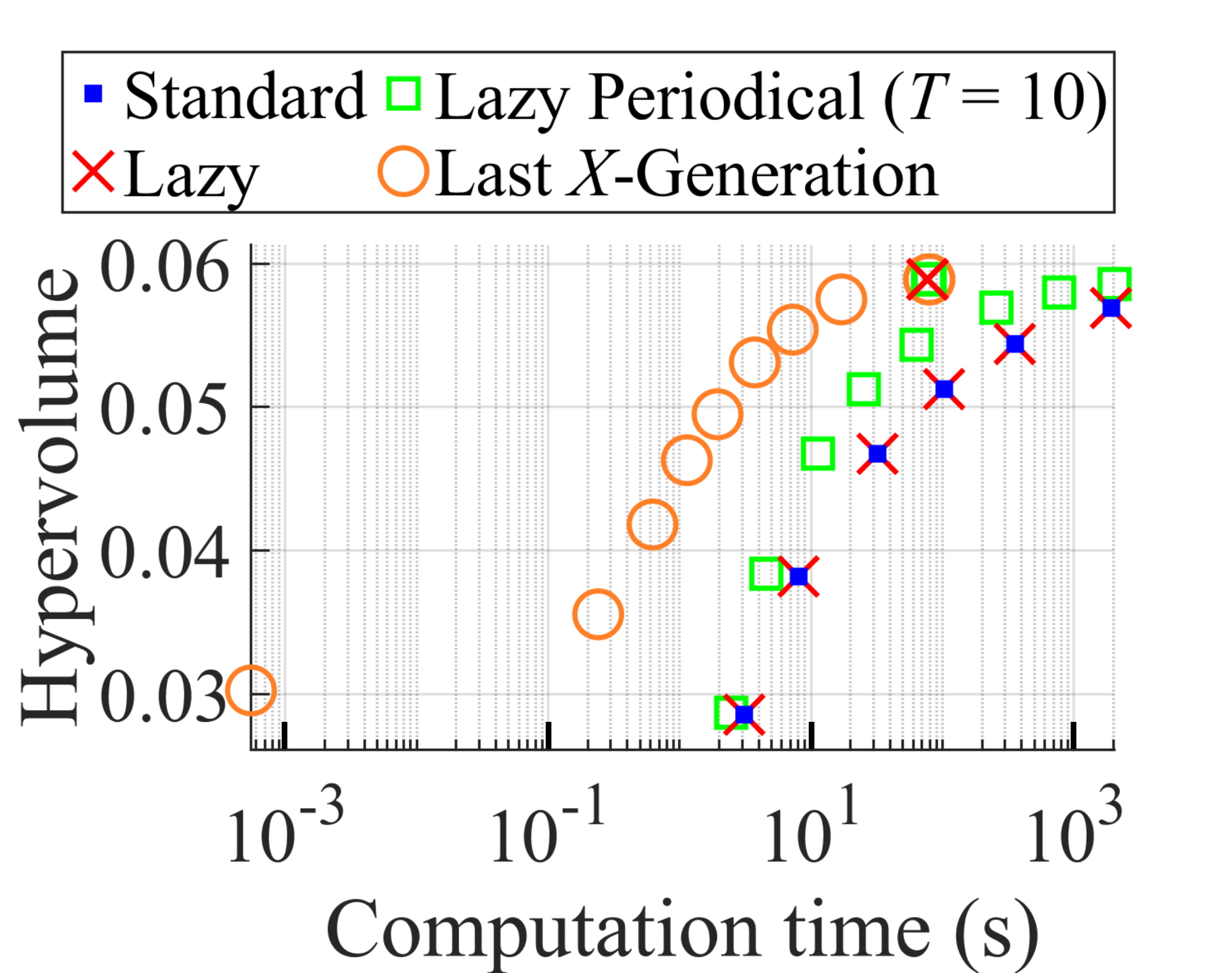}}
  \subfigure[Quality and memory size.] {\includegraphics[width=0.49\linewidth,trim=0 0 0 0,clip]{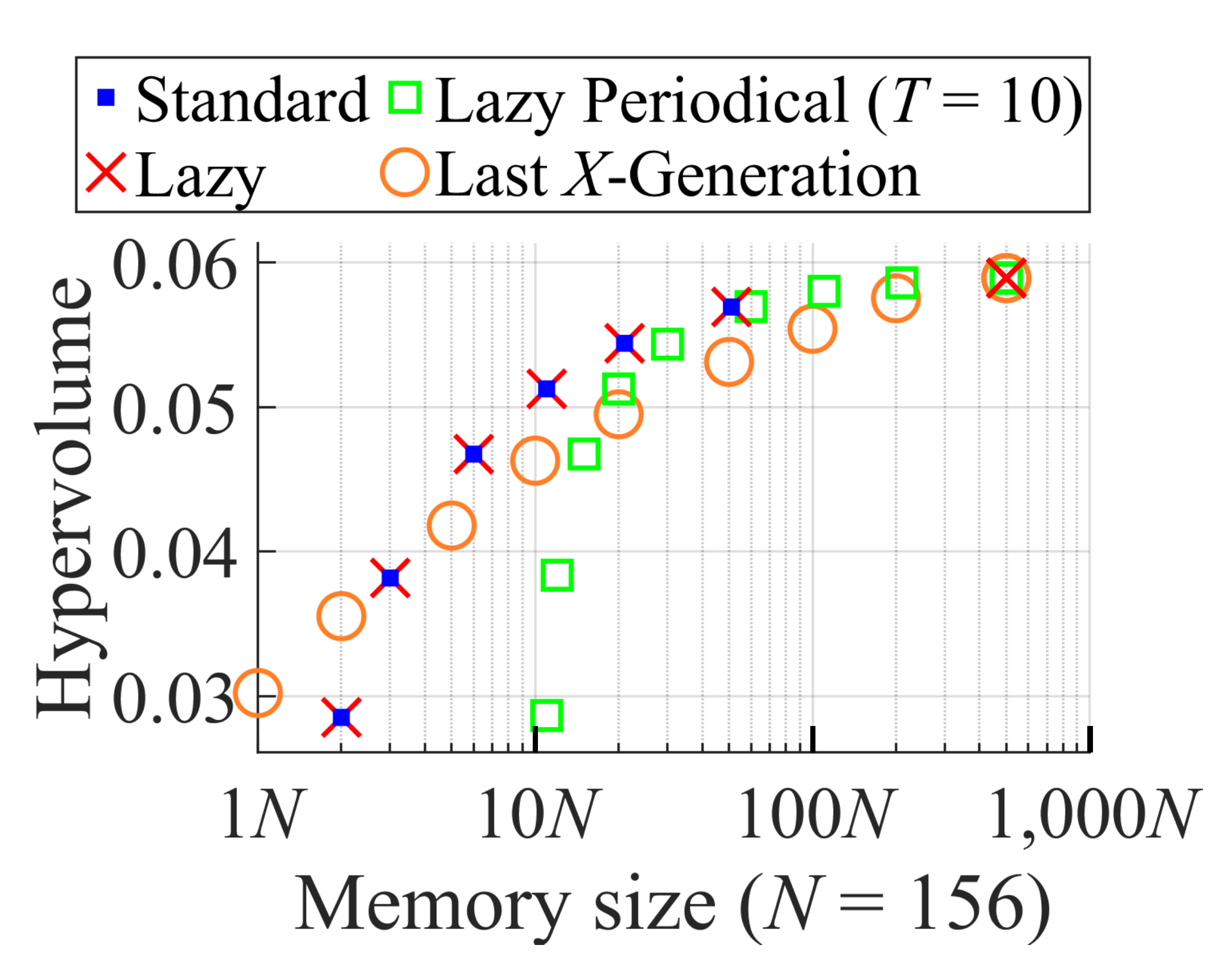}}
    \caption{Relation among the final solution set quality, the total computation time, and the required memory size in the results in Fig.~\ref{fig:hypervolume} (g) on 8-objective Minus-DTLZ2 with NSGA-II.}\label{fig:memory_and_time}
    \vspace{-1em}
 \end{figure}

To further examine the relation among the quality of the selected final solution set, the total computation time and the required memory size, all experimental results mentioned in Section~\ref{sec:Discussion_quality} (shown in the supplementary file) are redrawn on a quality-time plane and a quality-memory plane. As an example, the results on 8-objective Minus-DTLZ2 with NSGA-II in Fig.~\ref{fig:hypervolume} (g) are redrawn in Fig.~\ref{fig:memory_and_time} (a) with the quality-time plane and Fig.~\ref{fig:memory_and_time} (b) with the quality-memory plane. All results on the 72 combinations of the 3 EMO algorithms and the 24 test problems are included in the supplementary file.\par
Fig.~\ref{fig:memory_and_time} (a) clearly demonstrates the high time efficiency of the last $X$-generation strategy and the low time efficiency of the standard and lazy strategies. Fig.~\ref{fig:memory_and_time} (b) demonstrates the high memory efficiency of the standard and lazy strategies and the low memory efficiency of the last $X$-generation strategy. Compared with the standard and lazy strategies, the lazy periodical strategy with $T=10$ shows higher time efficiency in Fig.~\ref{fig:memory_and_time} (a) and lower memory efficiency in Fig.~\ref{fig:memory_and_time} (b). However, when the memory size is much larger than $10N$ (e.g., $100N$) in Fig.~\ref{fig:memory_and_time} (b), the lazy periodical strategy shows the best hypervolume performance since the standard and lazy strategies have no results due to their low time efficiency.\par
 When the available memory is enough to store all examined solutions, Fig.~\ref{fig:memory_and_time} (and all the other experimental results in this paper) suggests the use of an unbounded external archive, which leads to the best hypervolume performance of the selected solution set. When the available computation time is severely limited, it is a good idea to use the last $X$-generation strategy with a relatively large memory. Since the computation time of this strategy is mainly for the final subset selection, it is not difficult to choose an appropriate value of $X$ using the available computation time (since the number of candidate solutions directly depends on the value of $X$). When the available memory size is severely limited, the lazy strategy seems to be a good choice. However, since this strategy needs a long computation time even when the archive size is relatively small (e.g., $30N$), it may be a good idea to use the lazy periodical strategy at the cost of a slight increase of the memory size (i.e., memory for storing additional $TN$ solutions). Our experimental results also show that the final population can be better than the selected solution set in some cases when the archive size is very small. Those cases can be easily handled by choosing a better one between the final population and the selected solution set. 
\section{Conclusion}\label{sec:Conclusion}
In this paper, we examined the effects of the archive size on the standard archiving strategy where the archive is updated at every generation. Our experimental results showed that maintaining a medium-size archive is more time-consuming than maintaining a small-size archive or a huge-size archive (e.g., an unbounded archive). In order to decrease the computation time, we proposed an idea of updating the archive periodically (i.e., at every $T$ generations). We demonstrated that this idea clearly decreases the computation time with no deterioration of the final solution set quality at the cost of a slight increase of the memory requirement. We also proposed an idea of simply storing all solutions only in the last $X$ generations. We demonstrated that this idea can drastically decrease the computation time. Based on our experimental results, we obtained the following suggestions. When we have enough memory to store all examined solutions, it is a good idea to use an unbound archive and remove dominated solutions after the termination of an EMO algorithm (i.e., after all examined solutions are stored). If the computation time is severely limited, the last $X$-generation strategy can be used by appropriately choosing the value of $X$ based on the available computation time.\par
\textcolor{black}{Our experimental results also suggested some future research topics. One direction is the performance improvement of greedy subset selection algorithms with respect to the quality of the selected solution set. Another direction is to examine other efficient truncation algorithms (in addition to the distance-based approach in this paper). Since our experiments were performed on artificial test problems, the use of real-world problems for examining archiving strategies is an important future research topic.}

\bibliographystyle{IEEEtran}
\bibliography{reference}
\vfill

\end{document}